\documentclass[journal]{IEEEtran}
\usepackage{amsmath, amsfonts, amssymb, graphicx, makecell, url, cite, bm, stfloats, float, balance, stackengine}
\usepackage[caption=false,font=normalsize,labelfont=sf,textfont=sf]{subfig}
\usepackage[colorlinks,bookmarksopen,bookmarksnumbered,citecolor=red,urlcolor=red]{hyperref}

\newcommand{\bom}{\mbox{\boldmath $\omega$}}

\newcommand{\figref}[1]{Fig.~\ref{#1}}
\newcommand{\secref}[1]{Sec.~\ref{#1}}

\def\delequal{\mathrel{\ensurestackMath{\stackon[1pt]{=}{\scriptstyle\Delta}}}}

\begin{document}

\title{Adaptive Force-Based Control of Dynamic Legged Locomotion over Uneven Terrain}

\author{Mohsen Sombolestan and Quan Nguyen 
\thanks{M. Sombolestan and Q. Nguyen are with the Department of Aerospace and Mechanical Engineering, University of Southern California, Los Angeles, CA 90089, email: {\tt somboles@usc.edu, quann@usc.edu}.}
}

\maketitle
\begin{abstract}
Agile-legged robots have proven to be highly effective in navigating and performing tasks in complex and challenging environments, including disaster zones and industrial settings. However, these applications commonly require the capability of carrying heavy loads while maintaining dynamic motion. Therefore, this paper presents a novel methodology for incorporating adaptive control into a force-based control system. Recent advancements in the control of quadruped robots show that force control can effectively realize dynamic locomotion over rough terrain. By integrating adaptive control into the force-based controller, our proposed approach can maintain the advantages of the baseline framework while adapting to significant model uncertainties and unknown terrain impact models. Experimental validation was successfully conducted on the Unitree A1 robot. With our approach, the robot can carry heavy loads (up to 50\% of its weight) while performing dynamic gaits such as fast trotting and bounding across uneven terrains.
\end{abstract}

\section{Introduction}
\label{sec: introduction}
\IEEEPARstart{L}{egged} robots have numerous potential uses, from search and rescue operations to autonomous construction. To perform these tasks effectively, the robot must accurately understand the environment it will be operating in. However, due to the complexity of the robot and the environment, the robot's model might contain a significant level of uncertainty and affect the robot's stability, particularly when performing agile movements. To overcome these challenges, there is a need to develop a control framework that can effectively compensate for these uncertainties in real-time.

The utilization of convex model predictive control (MPC) with the single rigid body (SRB) model in legged robots \cite{DiCarlo2018} has greatly enhanced the real-time implementation of diverse walking gaits. Unlike the balance controller based on quadratic programming \cite{Focchi2017}, MPC offers the capability to perform agile motions like jumping \cite{Nguyen2019, Park2015} and high-speed bounding \cite{Park2017} for quadruped robots. Additionally, MPC exhibits robustness when traversing rough and uneven terrains. However, it is important to note that MPC assumes perfect knowledge of the dynamic model.

To enhance trajectory tracking in the presence of unknown and changing disturbances, researchers have explored the combination of MPC with adaptive control techniques \cite{Fukushima2007, Adetola2009, Pereida2018}. Additionally, parameter estimation techniques have been employed to improve the robustness of the control system further \cite{Lu2021}. These approaches aim to adapt the controller and estimate system parameters to effectively compensate for uncertainties and disturbances, leading to improved trajectory tracking performance. It is worth noting that all of these studies were conducted using a position-based controller model.

In this work, we tackle the legged robot locomotion issue in real-world scenarios with significant uncertainty. The uncertainty can come from both the robot model and the environment. Since our proposed method is based on a force controller, it retains the advantage of robustness to uneven terrain. Thanks to MPC as our baseline controller, our framework can be extended to different locomotion gaits and trajectories without adjusting the controller parameters.
Moreover, in our control system, we effectively manage substantial model uncertainty by utilizing the adaptive controller. By implementing adaptive control, our framework evolves into a versatile solution for mitigating persistent disturbances across various operations and over time. Given the adaptive control's capability to address uncertainties continuously, it provides a practical approach for real-world applications in legged robot autonomy, such as rescue missions, inspections, and logistics. This ability to compensate for persistent disturbances in real-world scenarios eliminates the need for recalibration for various tasks, enabling a thorough online operation. As a result, this represents a key contribution to our work, offering a comprehensive approach for legged robot applications and facilitating movement across diverse terrains with unknown impact models.

\begin{figure}[!t]
	\center
    \includegraphics[width=1\linewidth]{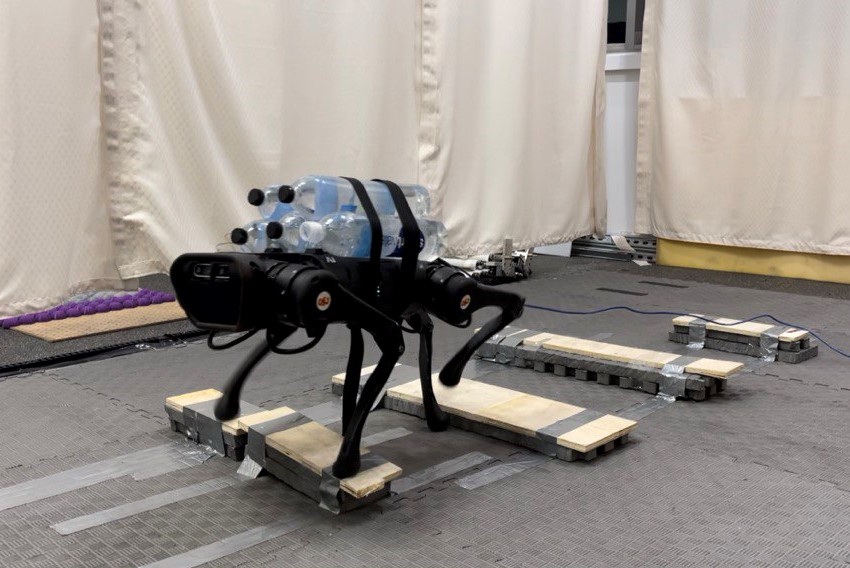}
	\caption{Our proposed adaptive MPC is successfully validated in an experiment on a Unitree A1 robot while carrying an unknown load of 5 kg (almost 50\% of body weight) on rough terrain. Experimental results video: \protect\url{https://youtu.be/5t1mSh0q3lk}.} 
	\label{fig: first fig}
\end{figure}

\subsection{Related Works}

\subsubsection{Offline Learning}
The offline learner can leverage a model-based control approach or learn the control system from scratch. Using a model-based method, researchers mainly target learning the dynamic to improve the controller performance \cite{Nagabandi2020}. One example of this approach is integrating deep learning with MPC, in which the proposed model tries to learn the cost or dynamic terms of an MPC \cite{Amos2018}. This hybrid method shows considerable improvement for the aerial robot \cite{Chee2022} when learning the dynamic model from experimental data. 
The major limitation of this method is that it is restricted to the dynamic model learned during the training phase. However, the dynamic model is prone to frequent changes in real-world scenarios due to environmental uncertainties and external disturbances. 

There has been growing interest in utilizing reinforcement learning (RL) to train models from scratch to overcome the limitations of previous approaches. The key advantage of RL models is their ability to adapt swiftly to changes in real-world environments due to being trained in diverse environments with varying properties. In the case of quadruped robots, an RL model can directly predict appropriate joint torques for traversing different types of terrain, as demonstrated by Chen et al. \cite{Chen2022}. Additionally, by training the model to learn foot positions, Bellegarda et al. \cite{Bellegarda2021} enable quadrupeds to run quickly while carrying unknown loads. However, these methods heavily rely on domain randomization during training to generalize to challenging environments. Yang et al. \cite{Yang2022} also propose an end-to-end RL method that utilizes proprioceptive states and visual feedback to predict environmental changes.

\vspace{2pt}

\subsubsection{Online Learning}
To address inaccuracies in model-based controllers, researchers have explored an alternative approach using online learning, mainly supervised learning methods \cite{Shalev-Shwartz2011, Bottou2003, Nagabandi2019}. In this approach, the focus is on learning disturbances online \cite{Duong2022}, and in some cases, researchers also aim to learn the dynamics of the system itself \cite{Jiahao2023OnlineRobots}. Furthermore, this approach has been successfully applied for online calibration of kinematic parameters in legged robots \cite{Yang2022a}. In addition, a recent study has developed a Lipschitz network method to bridge the model-reality gap in real-time \cite{Zhou2022}. The online learning method is closely related to adaptive control, and numerous studies have explored combining these two approaches \cite{Annaswamy2021}. This combination aims to leverage the advantages of both methods, allowing for dynamic adaptation and continuous learning from real-time data to improve control system performance. Perhaps closest to our work in terms of online adaption is the learning method presented in \cite{Sun2021} for legged robots. The authors correct the model behind the controller using a supervised learner while the robot is walking in an unknown environment. The data is collected during the robot's operation to learn a linear residual model that can compensate for system errors. However, in the transition from simulation to experiment, the acceleration estimators make noisy data required for training the model. As a result, the method is only applied to estimate the linear terms since the angular terms data proved to be too noisy to be helpful in the model.

To enhance controller efficiency and performance, auto-tuning methods, particularly for PID controllers, have gained widespread use \cite{Zhuang1993AutomaticControllers, Astrom1993AutomaticSurvey}. These methods fall into two categories: model-based and model-free. Model-based approaches use system model information, often employing the gradient of the performance criterion to enhance local performance \cite{Kumar2021DiffLoop:Loop}. In contrast, model-free methods, such as Markov chain Monte Carlo \cite{Loquercio2022AutoTune:Flight}, Gaussian process \cite{Calandra2014BayesianLocomotion, Lizotte2007AutomaticRegression}, and deep neural network \cite{Mehndiratta2021CanControllers}, approximate gradients or surrogate models to boost performance. However, model-based approaches may fail in real-world scenarios due to imperfect dynamic knowledge, and model-free methods like Bayesian optimization can be inefficient in high-dimensional parameter tuning.
Addressing this, recent works \cite{Cheng2023DiffTune+:Auto-Differentiation, Cheng2023DiffTune:Auto-Differentiation} directly obtain the gradient of the loss function with respect to controller parameters and apply it to gradient descent for performance improvement. Auto-tuning generally requires data or a fixed model for training and, therefore, does not fit well for real-time and fast adaptation to significant model uncertainty.

\vspace{5pt}

\subsubsection{Adaptive Control}
Adaptive control aims to tune the controller's variables online during deployment \cite{Astrom1991}. Adaptive control has been applied for manipulation tasks to robotic arms \cite{Slotine1987}, mobile robots \cite{Liu1998, Li2008, Culbertson2021}, and quadruped robots \cite{Sombolestan2022, Sombolestan2023}. The conventional Model Reference Adaptive Control (MRAC) architecture was initially designed for controlling linear systems in the presence of parametric uncertainties \cite{Slotine1991, Zhang2017}. However, it cannot characterize the input/output performance of the system during the transient phase. To address this limitation and improve the transient performance of adaptive controllers, the $L_1$ adaptive control offers several advantages over traditional MRAC, such as decoupling adaptation and robustness within a control framework \cite{L1_adaptive}. In addition, incorporating a low-pass filter in adaptation law allows the $L_1$ adaptive control to provide stability \cite{Cao2007} and transient performance \cite{Cao2006}. Therefore, the $L_1$ adaptive control technique guarantees robustness with fast adaptation \cite{Hovakimyan2010}, an essential criterion in dynamic robotics applications. Recently, by integrating $L_1$ adaptive controller and Bayesian learner, researchers leverage the fast adaption performance of the $L_1$ adaptive controllers and introduce a safe simultaneous control and learning framework \cite{Gahlawat2020, Gahlawat2021}. 

For legged robots, the adaptive controller has also been employed to find the value and location of the center of mass \cite{Tournois2017}. Our work on $L_1$ adaptive control for bipedal robots \cite{Nguyen2015} considers a Control Lyapunov Function (CLF)-based controller as a closed-loop nonlinear reference model for the $L_1$ adaptive controller. It was validated for the robot's walking \cite{Nguyen2018} and running \cite{Sreenath2013}. However, the control framework in this prior work is based on Hybrid Zero Dynamics \cite{Grizzle2008}, which uses joint position control to track the desired trajectory from optimization for each robot joint. Moreover, in \cite{Minniti2021}, an adaptive control based on a CLF is designed for quadrupeds to interact with unknown objects. Then, they combined the criteria derived by adaptive control as a constraint in an MPC framework. However, adding more inequality constraints to MPC makes the controller more complex in terms of computation. In our approach, we compute a residual vector to compensate for dynamic uncertainty, which makes the controller more time-efficient. Additionally, by employing our method, the robot can adapt to terrains with unknown impact models.

\subsection{Contributions}
A preliminary version of this research appeared in \cite{Sombolestan2021}; however, this paper presents several novel contributions to the prior work. This work incorporates the $L_1$ adaptive controller into the model predictive control (MPC). The proposed control system leverages MPC due to its robustness to uneven terrain, contact constraints, and generalization to different locomotion gaits. Moreover, by integrating adaptive control into MPC, the proposed model can compensate for significant model uncertainty. In the previous work \cite{Sombolestan2021}, the robot can only perform quasi-static walking; however, in this work, the robot can perform dynamic motions thanks to MPC. Finally, the authors present new hardware experiments to demonstrate the effectiveness of the proposed adaptive MPC (as illustrated in \figref{fig: first fig}). The main contributions of the paper are as follows:
\begin{itemize}
\item We introduce a novel control system that combines the $L_1$ adaptive control into the force-based control system, designed to address the challenges posed by model uncertainty in real-world applications. 

\item Thanks to MPC, our approach offers greater versatility as it can be adapted to a wide range of locomotion gaits and trajectories. Moreover, our method can handle terrain uncertainty, allowing the robot to navigate rough terrains and high-sloped terrain, such as grass and gravel.

\item By integrating the adaptive control into MPC, it is possible for quadruped robots to carry an unknown heavy load (up to 50\% of the robot's weight) across challenging terrains, with the capability of executing dynamic gaits such as fast trotting and bounding. This is a significant improvement compared to our previous work, which only allowed the robot to perform quasi-static walking.

\item The combination of using MPC for both the reference model and the real model in the adaptive controller makes the control system computationally expensive, leading to potential delays in computation. To ensure real-time performance, we have developed an update frequency scheme for the control system, which allows for the optimized allocation of processing resources to each control component. 

\item  Our proposed approach enables the control system to adapt to terrains with unknown impact models, such as soft terrain. Traversing soft terrain is a challenging task for quadruped robots. Using our method, the A1 robot can walk on double-foam terrain in different directions. In comparison, the robot cannot maintain its balance using the baseline controller, resulting in a collapse.

\end{itemize} 

The remainder of the paper is organized as follows. \secref{sec: background} presents the baseline control architecture for quadruped robots and provides some knowledge on force-based controllers. In \secref{sec: control overview}, we will briefly present an overview of our control approach. Then, our proposed adaptive force-based controller using balance controller and MPC will be elaborated in \secref{sec: adaptive control} and \secref{sec: adaptive MPC}, respectively. Furthermore, the numerical and experimental validation are shown in \secref{sec: Results}. Finally, \secref{sec: conclusion} provides concluding remarks.  
\section{Preliminaries} \label{sec: background}
In this section, we present the background on the control architecture of quadruped robots and describe each control component. According to \cite{Bledt2018}, the robot’s control system consists of several modules, including a high-level controller, low-level controller, state estimation, and gait scheduler as presented in \figref{fig: ControlOverview}. 

A reference trajectory can be generated for high-level control from user input and state estimation. The gait scheduler defines the gait timing and sequence to switch between each leg’s swing and stance phases. The high-level part controls the position of the swing legs and optimal ground reaction force for stance legs based on the user commands and gait timing. As the baseline for the stance leg controller, we will use two common approaches: 1) quadratic program (QP) based balancing controller \cite{Focchi2017} and 2) model predictive control (MPC) \cite{DiCarlo2018}. The low-level leg control converts the command generated by high-level control into joint torques for each motor. These modules of the control architecture will be described briefly in the following subsections. More details can be found in \cite{Bledt2018, DiCarlo2018, Focchi2017}.

\subsection{Gait Scheduler}
The A1’s gait is defined by a finite state machine using a leg-independent phase variable to schedule contact and swing phases for each leg \cite{Bledt2018}. The gait scheduler utilizes independent boolean variables to define contact states scheduled $\bm{s}_{\phi} \in \{1 = contact, 0 = swing\}$ and switch each leg between swing and stance phases. Based on the contact schedule, the controller will execute either position control during swing or force control during stance for each leg.

In our previous work \cite{Sombolestan2021}, we focus on the application of load-carrying tasks, where the load is unknown to the robot or the control system. Having more legs on the ground during walking could also mean that the robot could produce a more significant total ground reaction force to support the heavy load. Therefore, we used a quasi-static walking gait to maximize the number of legs on the ground during walking (i.e., three stance legs and one swing leg throughout the gait). However, in this paper, our framework is not limited by any specific gait. Similar to the baseline MPC control approach \cite{DiCarlo2018}, the approach can work for different gaits by only changing the gait definition in the gait scheduler. 

\subsection{Desired Trajectory}
The desired trajectory is generated based on the robot’s velocity command. The robot operator commands $xy$-velocity and yaw rate, and then $xy$-position and yaw are determined by integrating the corresponding velocity. $z$ position contains a constant value of $0.3 m$, and the remaining states (roll, roll rate, pitch, pitch rate, and $z$-velocity) are always zero.

\begin{figure}[t!]
	\centering
	\includegraphics[width=1\linewidth]{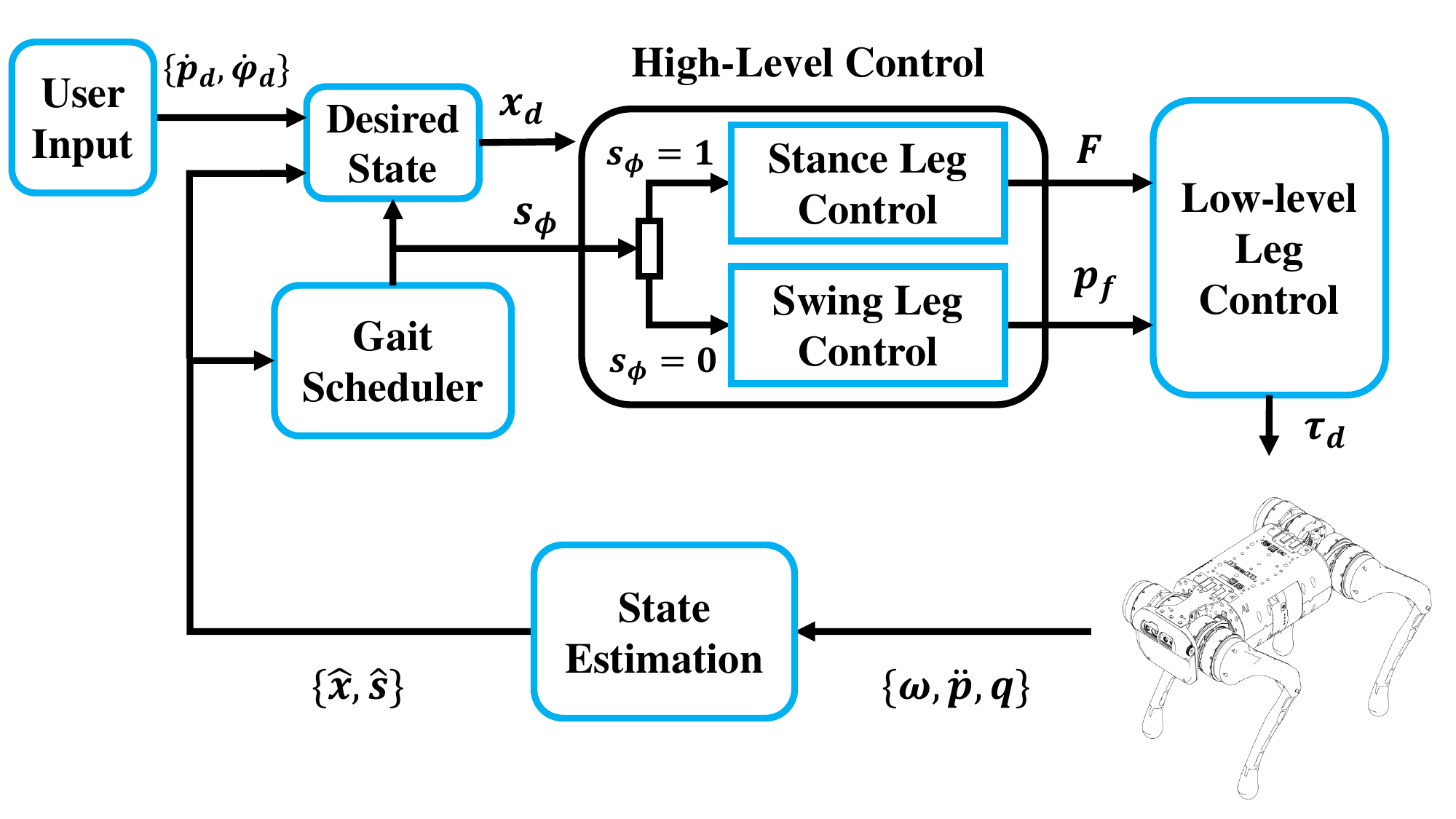}
	\caption{{\bfseries Baseline Control Structure.} Block diagram of a control architecture for a quadruped robot. For the stance leg control, we use two common baseline control systems: QP-based balancing controller and MPC.}
	\label{fig: ControlOverview}
\end{figure}

\subsection{Single Rigid Body (SRB) Model of Robot} \label{sec: simplified robot dynamic}
Due to the complexity of the legged robot, a simplified rigid-body model has been used to present the system’s dynamic. This model lets us calculate the ground reaction forces (GRFs) in real-time. A few  assumptions have been made to achieve simplified robot dynamics\cite{DiCarlo2018}:

\textit{Assumption 1:} The robot has low inertia legs, so their effect is negligible on the robot’s rigid body dynamic.

\textit{Assumption 2:} For small values of roll ($\phi$) and pitch ($\theta$), the rotation matrix $\bm{R}$ which transforms from the body to world coordinates, can be approximated as the rotation matrix corresponding to the yaw angle ($\psi$):
\begin{align}\label{eq:rotation matrix}
\bm{R} \cong \bm{R}_z(\psi) = \left[\begin{array}{ccc}  
\cos(\psi) & -\sin(\psi) & 0 \\  
\sin(\psi) & \cos(\psi) & 0  \\
0         &      0     & 1
\end{array} \right] 
\end{align}
Therefore, by defining the robot’s orientation as a vector of Z-Y-X Euler angles $\bm{\Theta} = [\phi, \theta, \psi]^T$, the rate of change of the robot’s orientation can be approximated as \cite{DiCarlo2018}:
\begin{align}\label{eq:rpy rate of change}
\dot{\bm{\Theta}} \cong \bm{R}_z(\psi) \bm{\omega}_b
\end{align}
where $\bm{\omega}_b$ is the robot’s angular velocity in the world frame.

\textit{Assumption 3:} For small angular velocity, the following approximation can be made: 
\begin{align}
\frac{d}{dt}(\bm{I}_G \bom_b) = \bm{I}_G \dot \bom_b + \bom_b \times (\bm{I}_G \bom_b) \approx \bm{I}_G \dot \bom_b
\end{align}
where $\bm{I}_G\in \mathbb{R}^{3 \times 3}$ is the moment of inertia in the world frame.

Based on the above assumptions, the state representation of the system is as follows \cite{DiCarlo2018}:
\begin{align}\label{eq:SS}
\left[\begin{array}{c}  
    \dot{\bm{p}}_{c} \\ 
    \bm{\dot{\Theta}} \\ 
    \ddot{\bm{p}}_{c} \\  
    \dot \bom_{b} 
\end{array} \right] = & 
\underbrace{\left[\begin{array}{cccc} 
    \bm{0}_3 & \bm{0}_3 & \bm{1}_3 & \bm{0}_3 \\ 
    \bm{0}_3 & \bm{0}_3 & \bm{0}_3 & \bm{R}_z(\psi)\\ 
    \bm{0}_3 & \bm{0}_3 & \bm{0}_3 & \bm{0}_3 \\ 
    \bm{0}_3 & \bm{0}_3 & \bm{0}_3 & \bm{0}_3
\end{array} \right]}_{\bm{D} \in \mathbb{R}^{12 \times 12} }
\underbrace{\left[\begin{array}{c}  
    {\bm{p}}_{c} \\ 
    \bm{\Theta} \\ 
    \dot{\bm{p}}_{c} \\ 
    \bom_{b} 
\end{array} \right]}_{\bm{X} \in \mathbb{R}^{12} } + \\ \nonumber
& \underbrace{\left[\begin{array}{c}
    \bm{0}_{6 \times 12} \\
    \bm{M}^{-1} \bm{A} 
\end{array} \right]}_{\bm{H} \in \mathbb{R}^{12 \times 12} } \bm{F} + \left[\begin{array}{c}
    \bm{0}_{6 \times 1} \\
    \bm{G} 
\end{array} \right]
\end{align}
with 
\begin{align}\label{eq: ss parameters}
&\bm{M} = \left[\begin{array}{cc} m \bf{1}_3 & \bm{0}_3 \\ \bm{0}_3 & \bm{I}_G \end{array} \right] \in \mathbb{R}^{6 \times 6} \\[1ex] \nonumber
&\bm{A} = \left[\begin{array}{ccc} \bf{1}_{3} & \dots & \bf{1}_{3}  \\ \hspace{-0.1cm}[\bm{p}_{1} - \bm{p}_{c}] \times & \dots & [\bm{p}_{4} - \bm{p}_{c}] \times \end{array} \right] \in \mathbb{R}^{6 \times 12} \\[1ex] \nonumber
&\bm{G} = \left[\begin{array}{c} \bm{g} \\ \bm{0}_{3 \times 1} \end{array} \right] \in \mathbb{R}^{6}
\end{align}
where $m$ is the robot’s mass, $\bm{g}\in \mathbb{R}^{3}$ is the gravity vector, $\bm{p}_{c}\in \mathbb{R}^{3}$ is the position of the center of mass (COM), $\bm{p}_{i}\in \mathbb{R}^{3}$ ($i \in \{1,2,3,4\}$) are the positions of the feet, $\ddot{\bm{p}}_{c}\in \mathbb{R}^{3}$ is body’s linear acceleration, $\dot \bom_b\in \mathbb{R}^{3}$ is angular acceleration, and $\bm{F} = [\bm{F}_1^T, \bm{F}_2^T, \bm{F}_3^T, \bm{F}_4^T]^T \in \mathbb{R}^{12}$ are the ground reaction forces acting on each of the robot’s four feet. The term $[\bm{p}_{i} - \bm{p}_{c}] \times$ is the skew-symmetric matrix representing the cross product $(\bm{p}_{i} - \bm{p}_{c}) \times  \bm{F}_i$. Note that $\bm{p}_{i}$ and $\bm{F}_i$ are presented in the world frame. Therefore, the state representation of the system can be rewritten in the compact form:
\begin{align}\label{eq: SS}
 \bm{\dot{X}} = \bm{D} \bm{X} + \bm{H} \bm{F}  + \left[\begin{array}{c}
    \bm{0}_{6 \times 1} \\
    \bm{G} 
\end{array} \right]
\end{align}

\subsection{Balance Controller} \label{sec: balance controller}
One of the baseline control approaches for calculating GRFs for quadruped robots is the balance controller presented in \cite{Focchi2017} based on a quadratic program (QP) solver. Based on the assumptions presented in \secref{sec: simplified robot dynamic}, the approximated dynamic model between the body acceleration and GRFs is as follows:
\begin{align}
\label{eq:linear_model}
\underbrace{\left[\begin{array}{ccc} \bf{1}_{3} & \dots & \bf{1}_{3}  \\ \hspace{-0.1cm}[\bm{p}_{1} - \bm{p}_{c}] \times & \dots & [\bm{p}_{4} - \bm{p}_{c}] \times \end{array} \right]}_{\bm{A}\in \mathbb{R}^{6 \times 12}} \bm{F} = \underbrace{\left[\begin{array}{c} m (\ddot{\bm{p}}_{c} +\bm{g}) \\ \bm{I}_G \dot \bom_b \end{array} \right]}_{\bm{b} \in \mathbb{R}^{6}}
\end{align}
and the vector $\bm{b}$ in \eqref{eq:linear_model} can be rewritten as:
\begin{align}\label{eq:b}
\bm{b} = \bm{M} (\left[\begin{array}{c}  \ddot{\bm{p}}_{c} \\  \dot \bom_{b} \end{array} \right] +  \bm{G}).
\end{align}
Since the model \eqref{eq:linear_model} is linear, the controller can naturally be formulated as the following QP problem \cite{Gehring2013}, which can be solved in real-time at $1~kHz$: 
\begin{align}\label{eq: BalanceControlQP}
\nonumber
\bm{F}^* =  \underset{\bm{F} \in \mathbb{R}^{12}}{\operatorname{argmin}}   \:\:  &(\bm{A} \bm{F} - \bm{b}_d)^T \bm{S} (\bm{A} \bm{F} - \bm{b}_d)  \\  & + \gamma_1 \| \bm{F} \|^2 + \gamma_2 \| \bm{F} - \bm{F}_{\textrm{prev}}^* \|^2\\
\mbox{s.t. }& \quad \:\: \underline{\bm{d}} \leq \bm{C} \bm{F} \leq \bar{\bm{d}} \nonumber\\
& \quad \:\: \bm{F}_{swing}^z=0 \nonumber
\end{align}
where $\bm{b}_d$ is the desired dynamics. The idea of designing $\bm{b}_d$ will be elaborated in \secref{sec: closed_loop}. The cost function in \eqref{eq: BalanceControlQP} includes terms that consider three goals, including (1) driving the COM position and orientation to the desired trajectories, (2) minimizing the force commands, and (3) minimizing the change of the current solution $\bm{F}^*$ with respect to the solution from the previous time-step, $\bm{F}^*_{prev}$. 
The priority of each goal in the cost function is defined by the weight parameters $\bm{S}\in \mathbb{R}^{6 \times 6}$, $\gamma_1$, $\gamma_2$ respectively.

The constraints in the QP formulation enforce friction constraints, input saturation, and contact constraints. 
The constraint $\underline{\bm{d}} \leq \bm{C} \bm{F} \leq \bar{\bm{d}}$ ensures that the optimized forces lie inside the friction pyramid and the normal forces stay within a feasible range. More details can be found in \cite{Focchi2017}.
Besides the friction constraint, we will enforce the force constraints for the swing legs, $\bm{F}_{swing}=\bm{0}$. The swing legs are then kept in the posing position until they switch to the stance phase. More details on swing leg control are provided in \secref{sec: swing leg}.

\subsection{SRB-based Convex MPC} \label{sec: mpc}
The calculation of GRFs in quadruped robots is often approached through Model Predictive Control (MPC) \cite{DiCarlo2018}. This method determines the optimal sequence of inputs over a finite-time horizon, taking into account any constraints within the dynamic model. Every time MPC is executed in the control system, only the first computed control input from the MPC cycle is applied. The inputs determined over the finite time horizon are only used for the optimization problem and are not directly applied in the control system.

To have the dynamic equation in the convenient state-space form, gravity should be added to the state. So, the system can represent as:
\begin{align}\label{eq: convenient SS}
 \bm{\dot{X}}^{c} = \bm{D}^c \bm{X}^c + \bm{H}^c \bm{F}   
\end{align}
where 

\begin{align}\label{eq: convenient SS components}
&\bm{X}^c = \left[\begin{array}{c} 
    \bm{p}_{c} \\ 
    \bm{\Theta} \\ 
    \dot{\bm{p}}_{c} \\ 
    \bom_{b}\\
    ||\bm{g}||
    \end{array} \right] \in \mathbb{R}^{13}
\\ \nonumber
&\bm{D}^c = \left[\begin{array}{ccccc} 
    \bm{0}_3 & \bm{0}_3 & \bm{1}_3 & \bm{0}_3 & \bm{0}_{3 \times 1}\\ 
    \bm{0}_3 & \bm{0}_3 & \bm{0}_3 & \bm{R}_z(\psi) & \bm{0}_{3 \times 1}\\ 
    \bm{0}_3 & \bm{0}_3 & \bm{0}_3 & \bm{0}_3 & \frac{\bm{g}}{||\bm{g}||}\\ 
    \bm{0}_3 & \bm{0}_3 & \bm{0}_3 & \bm{0}_3 & \bm{0}_{3 \times 1}\\
    \bm{0}_{1 \times 3} & \bm{0}_{1 \times 3} & \bm{0}_{1 \times 3} & \bm{0}_{1 \times 3} & 0
\end{array} \right] \in \mathbb{R}^{13 \times 13} \\ \nonumber
&\bm{H}^c = \left[\begin{array}{c}
    \bm{0}_{6 \times 12} \\
    \bm{M}^{-1} \bm{A} \\
    \bm{0}_{1 \times 12}
\end{array} \right] \in \mathbb{R}^{13 \times 12}
\end{align}

We consider a linear MPC problem with horizon length $k$ as follows:
\begin{align}
\label{eq:mpc}
\min_{\bm{F}_i} \quad & \sum_{i=0}^{k-1} {\bm{e}_{i+1}}^T \bm{Q}_i {\bm{e}_{i+1}} + {\bm{F}_{i}}^T \bm{R}_i \bm{F}_{i} \\ \nonumber
\textrm{s.t.} 
\quad & \bm{X}^{c}_{i+1} = \bm{D}_{t,i} \bm{X}^{c}_{i} + \bm{H}_{t,i} \bm{F}_{i} \\ \nonumber
& \underline{\bm{d}} \leq \bm{C} \bm{F}_{i} \leq \bar{\bm{d}}
\end{align}
where $\bm{F}_{i}$ is the computed ground reaction forces at time step $i$, $\bm{Q}_i$ and $\bm{R}_i$ are diagonal positive semi-definite matrices, $\bm{D}_{t,i}$ and $\bm{H}_{t,i}$ are discrete time system dynamics matrices. The ${\bm{e}_{i+1}}$ is the system state error at time step $i$ define as $\bm{e} = [\bm{e}_p,~\dot{\bm{e}}_p]^T \in \mathbb{R}^{12}$, with
\begin{align}\label{eq: state error}
\bm{e}_p = \left[\begin{array}{c} \bm{p}_{c}-\bm{p}_{c,d} \\ \log(\bm{R}_d \bm{R}^T)	\end{array} \right]\in \mathbb{R}^{6}, \quad 
\dot{\bm{e}}_p = \left[\begin{array}{c} \dot{\bm{p}}_{c}-\dot{\bm{p}}_{c,d} \\ \bom_b -\bom_{b,d} \end{array} \right]\in \mathbb{R}^{6},
\end{align} 
where $\bm{p}_{c,d}\in \mathbb{R}^{3}$ is the desired position of COM, $\dot{\bm{p}}_{c,d}\in \mathbb{R}^{3}$ is the desired body’s linear velocity, and $\bom_{b,d}\in \mathbb{R}^{3}$ is the desired body’s angular velocity. The desired and actual body orientations are described using rotation matrices $\bm{R}_d\in \mathbb{R}^{3 \times 3}$ and $\bm{R}\in \mathbb{R}^{3 \times 3}$, respectively. The orientation error is obtained using the exponential map representation of rotations \cite{Bullo1995, Murray2017}, where the $log(.):\mathbb{R}^{3 \times 3} \to \mathbb{R}^{3} $ is a mapping from a rotation matrix to the associated rotation vector \cite{Focchi2017}.
The constraint $\underline{\bm{d}} \leq \bm{C} \bm{F}_i \leq \bar{\bm{d}}$ is equivalent to the constraint in equation \eqref{eq: BalanceControlQP} at time step $i$.

\subsection{Swing Leg Control} \label{sec: swing leg}
For the swing legs, the final footstep location for each leg is calculated from the corresponding hip location using a linear combination of Raibert heuristic \cite{Raibert1986} and a feedback term from the capture point formulation \cite{Pratt2006, Bledt2018}. The final footstep locations ($\bm{p}_{f,i}$) are projected on an assumed ground plane and are calculated by:
\begin{align}
\bm{p}_{f,i} = \bm{p}_{h,i} + \frac{{T}_{\bm{c}_{\phi}}}{2}\bm{\dot{p}}_{c,d} + \sqrt{\frac{\bm{z}_{0}}{\|\bm{g}\|}}(\bm{\dot{p}}_{c} - \bm{\dot{p}}_{c,d}) 
\end{align}
where ${T}_{\bm{c}_{\phi}}$ is the stance time scheduled, $\bm{z}_{0}$ is the height of locomotion and $\bm{p}_{h,i}\in \mathbb{R}^{3}$ is the position of the corresponding hip $i$. A Beizer curve calculates the desired swing trajectory (including desired position $\bm{p}_{d,i}$ and velocity $\bm{v}_{d,i}$) for swing legs which starts from the initial lift-off position $\bm{p}_{0,i}$ and ends at the final touch-down location $\bm{p}_{f,i}$.

\subsection{Low-level Control} \label{sec: low-level control}
The low-level leg control can generate joint torque commands from the high-level controller. For low-level force control, the controller transforms the force vector to the hip frame by rotation matrix $\bm{R}$. Then, joint torques are calculated as follows:
\begin{align} \label{eq: torque_mapping}
\bm{\tau}_{stance, i} = -{\bm{J}(\bm{q}_i)}^{T} \bm{R}^{T}\bm{F}_{i}
\end{align}
where $\bm{J}(\bm{q}_i)\in \mathbb{R}^{3 \times 3}$ is the leg Jacobian matrix and $\bm{q_i}$ is the joints angle of leg $i$-th. 

To track the desired swing trajectory for each foot, a PD controller with a feedforward term is used to compute joint torques \cite{Bledt2018}:
\begin{align}
\bm{\tau}_{swing, i} = \bm{J}(\bm{q}_i)^{T}[\bm{K}_{p,p}(\bm{p}_{d,i} - \bm{p}_i)+\bm{K}_{d,p}(\bm{v}_{d,i}-\bm{v}_i)]
\end{align}
where $\bm{p}_{d,i}$ and $\bm{v}_{d,i}$ are desired foot position and velocity, respectively, $\bm{p}_i$ and $\bm{v}_i$ are actual foot position and velocity in the robot’s frame, $\bm{K}_{p,p}\in \mathbb{R}^{3 \times 3}$ and $\bm{K}_{d,p}\in \mathbb{R}^{3 \times 3}$ are the diagonal matrices of the proportional and derivative gains.

\section{Overview of the Proposed Approach} \label{sec: control overview}
This section will present an overview of our novel control architecture to incorporate adaptive control into the force control framework. While our approach is not limited to any specific adaptive control approach, we decided to use $L_1$ adaptive control \cite{Hovakimyan2010, Nguyen2015} thanks to its advancement in guaranteeing fast adaptation and smooth control signals. Note that our proposed control system is designed for the stance leg control part in the control architecture of the quadruped robot (see \figref{fig: ControlOverview}). 

Our prior work \cite{Nguyen2015} introduced an adaptive control based on Hybrid Zero Dynamics (HZD) \cite{Westervelt2003} for bipedal robots. HZD is a common control approach for bipedal robots since it can handle hybrid and underactuated dynamics associated with this kind of robot. In this paper, however, our approach leverages the combination of the adaptive control and force control system, which calculates ground reaction forces (GRFs) to achieve highly dynamic locomotion for quadrupeds \cite{Bledt2018}. The use of force control in legged robot systems has several key benefits, including increased robustness in the presence of challenging terrains \cite{Focchi2017} and the ability to accommodate a wide range of dynamic movements \cite{DiCarlo2018}, such as various types of locomotion gaits. By combining force control with adaptive control strategies that compensate for model uncertainty, achieving an enhanced control system with these advantages is possible.

\begin{figure}[t!]
        \centering
	\subfloat[Main structure]{\includegraphics[width=1.0\linewidth]{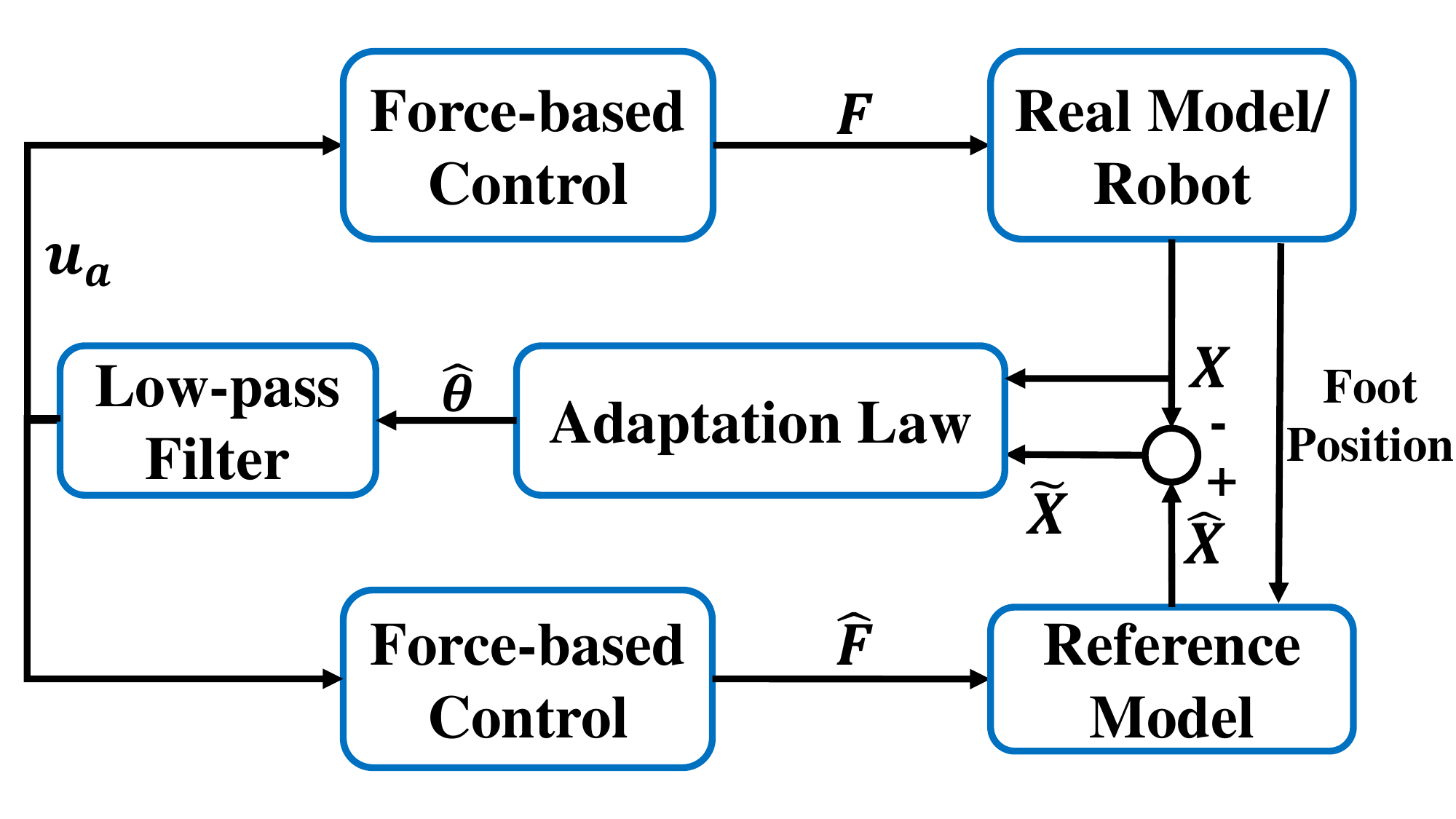}\label{fig: main adaptive structure}} \\
	\subfloat[Adaptive balance controller]{\includegraphics[width=1.0\linewidth]{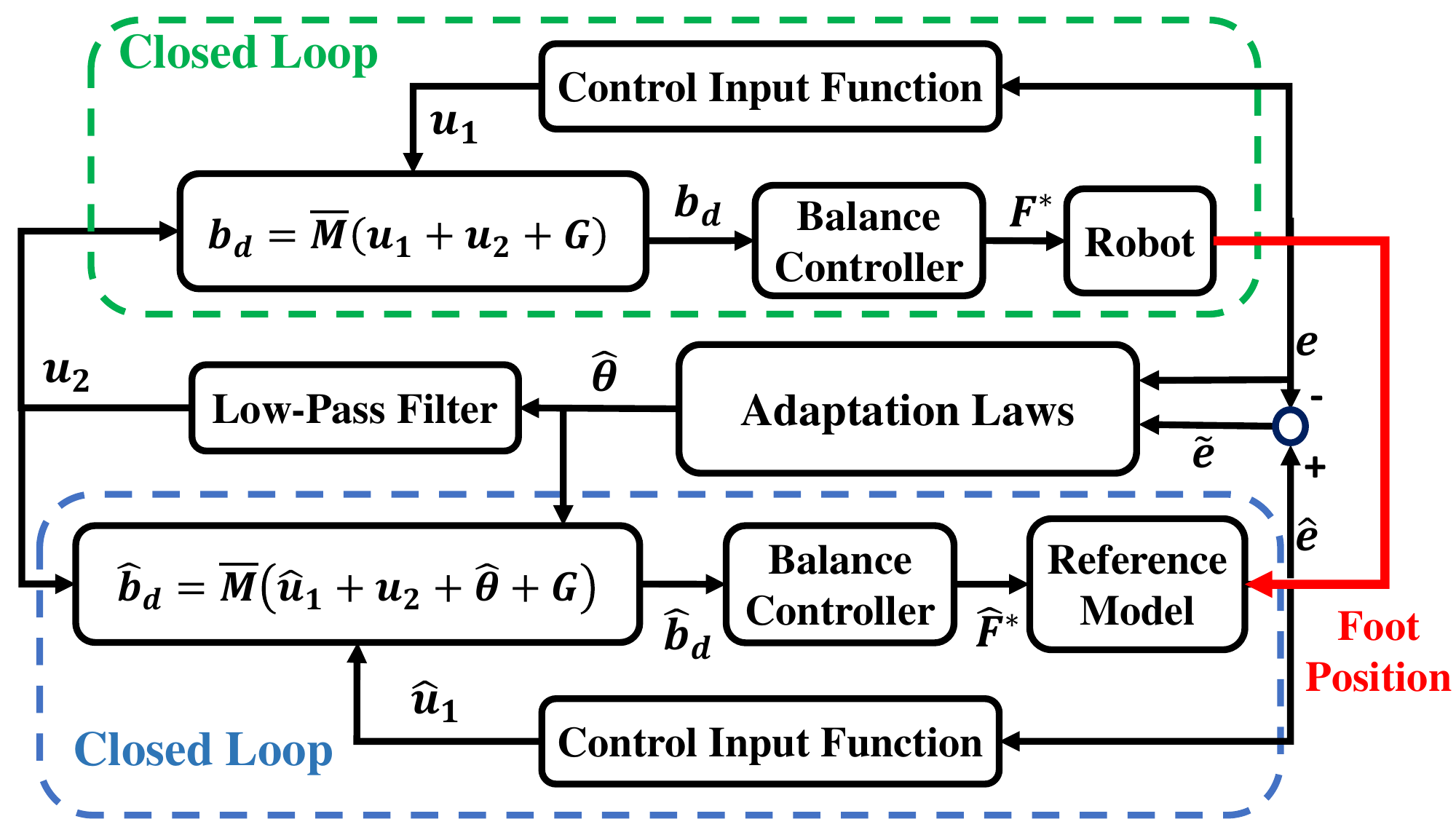}\label{fig: ControlDiagram_QP}} \\
	\subfloat[Adaptive MPC]{\includegraphics[width=1.0\linewidth]{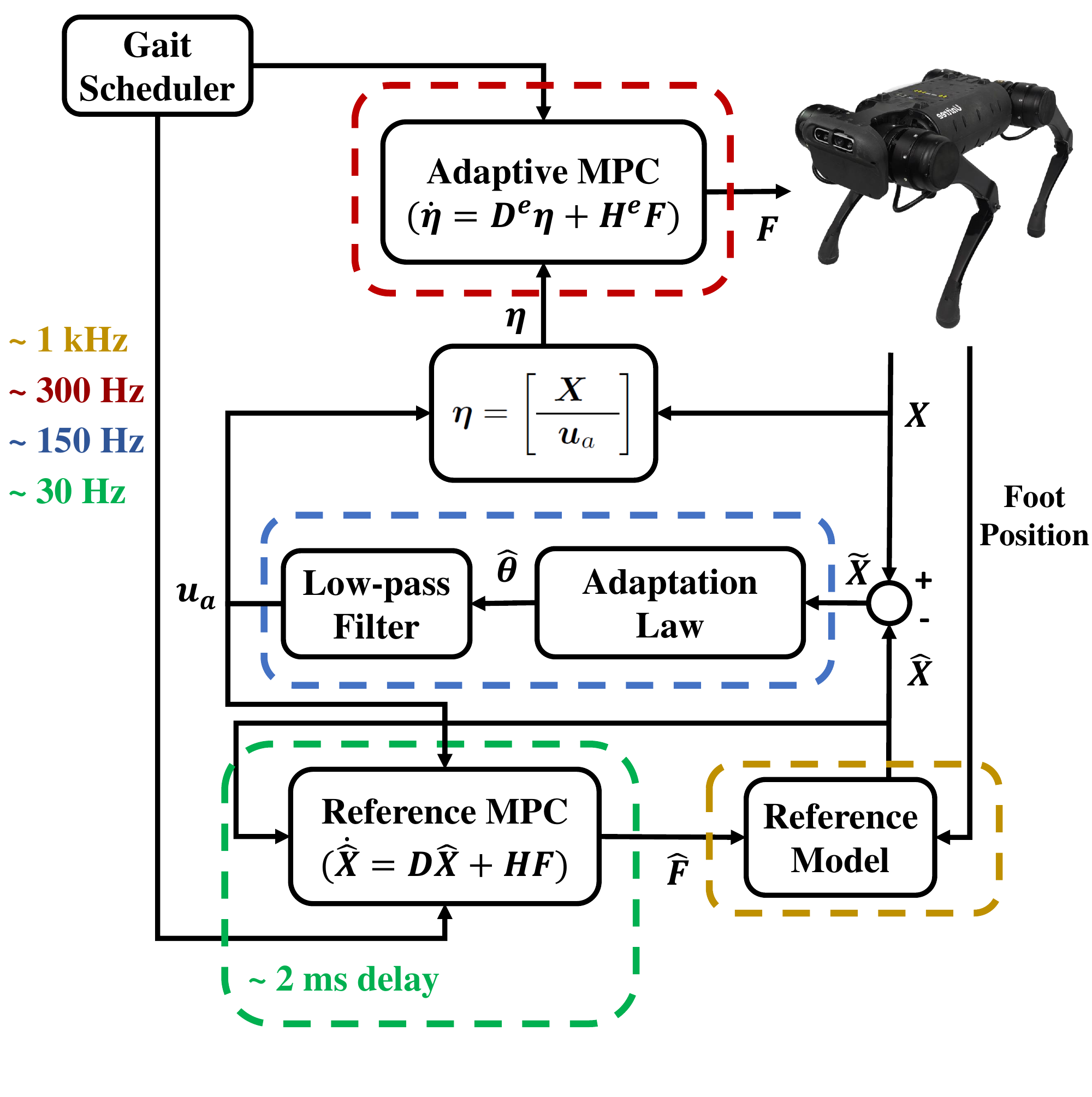}\label{fig: ControlDiagram_MPC}} 
    \caption{\textbf{Proposed adaptive force-based control system diagram.} a) The main structure of the proposed adaptive force-based control system, b) Block diagram of the proposed adaptive QP-based balancing controller, c) Block diagram of the proposed adaptive MPC. Each dashed line indicates the update frequency for  control components}
    \label{fig: adaptive framework}
\end{figure}

The overview of our proposed adaptive force-based control system is presented in \figref{fig: main adaptive structure}. By incorporating a $L_1$ adaptive controller, we aim to design a combined controller. The force-based controller calculates the optimal GRFs for following the desired trajectory. The adaptive controller calculates the residual parameters for compensating the nonlinear model uncertainty $\bm{\theta}$ in the system dynamic. Therefore, the goal is to adjust adaptive control signal $\bm{u}_a$ as well as adaptation law to estimate the model uncertainty ($\hat{\bm{\theta}}$) correctly and make the real model follows the reference model. For the reference model, we employ a similar linear model described in \eqref{eq: SS}, and we will update the reference model in real-time using an ODE solver. Moreover, the vector of uncertainties estimation $\hat{\bm{\theta}}$ typically has high frequency due to fast estimation in the adaptation law. Thus, we employ a low-pass filter to obtain smooth control signals. We use the same swing leg control to appropriately synchronize the reference and real models. This means that we also use the real model's foot position for the reference model. 

In the following sections, we will elaborate on integrating two different force-based controls as the baseline controller into the adaptive control. First, in \secref{sec: adaptive control}, we will describe the proposed method using a QP-based balancing controller, as presented in \figref{fig: ControlDiagram_QP}. Then, in \secref{sec: adaptive MPC}, we will show how to incorporate MPC into the adaptive controller in detail, as illustrated in \figref{fig: ControlDiagram_MPC}.

\section{Adaptive force-based control Using the Balance Controller}
\label{sec: adaptive control}
In this section, we use the balance controller as the force-based controller, previously demonstrated in \cite{Sombolestan2021}. In \secref{sec: adaptive MPC}, we will present our control framework for integrating the $L_1$ adaptive control into MPC.

\subsection{Closed-loop Dynamics} \label{sec: closed_loop}
The $L_1$ adaptive control is designed for trajectory tracking; however, the goal of the balance controller is to compute optimal GRFs. Hence, to integrate the balance controller presented in \secref{sec: balance controller} into $L_1$ adaptive control, we should relate the linear model described in \eqref{eq:linear_model} into the closed-loop dynamics.

Let us consider the system state error ($\bm{e}$) according to equation \eqref{eq: state error} as the state variable. Therefore, the closed-loop error dynamics in state-space form can be represented as follows:
\begin{align}
\label{fgBarDynamics}
\dot{\bm{e}} = \bm{D}_l \bm{e} + \bm{B}  \bm{u},
\end{align}
where
\begin{align}
\bm{D}_l = \left[ \begin{array}{cc} \bm{0}_6  & \bm{1}_6 \\ \bm{0}_6 & \bm{0}_6 \end{array} \right]\in \mathbb{R}^{12 \times 12}, \quad
\bm{B}  =  \left[ \begin{array}{c} \bm{0}_6 \\ \bm{1}_6 \end{array} \right] \in \mathbb{R}^{12 \times 6}
\end{align}
and $\bm{u}\in \mathbb{R}^{6}$ is the control input function. By employing a PD control law, we have
\begin{equation}\label{eq: PDcontrol}
\bm{u} = \begin{bmatrix}-\bm{K}_P & -\bm{K}_D\end{bmatrix} \bm{e},
\end{equation}
where $\bm{K}_P \in \mathbb{R}^{6 \times 6}$ and  $\bm{K}_D \in \mathbb{R}^{6 \times 6}$ are diagonal positive definite matrices. According to definition of matrices $\bm{D}_l$ and $\bm{B}$, from equation \eqref{fgBarDynamics} it can be obtained that:
\begin{align}\label{eq:epp}
\ddot{\bm{e}}_p = \left[\begin{array}{c} \ddot{\bm{p}}_{c} - \ddot{\bm{p}}_{c,d} \\ \dot \bom_{b} - \dot \bom_{b,d} \end{array} \right] = \bm{u},
\end{align}
where $\ddot{\bm{e}}_p$ is the derivative of $\dot{\bm{e}}_p$ presented in \eqref{eq: state error}, $\ddot{\bm{p}}_{c,d}$ and $\dot \bom_{b,d}$ are the desired COM linear acceleration and the desired angular acceleration, respectively. Since the desired trajectory is obtained from the velocity command, both desired accelerations $\ddot{\bm{p}}_{c,d}$ and $\dot \bom_{b,d}$ are zero vectors. Then from \eqref{eq:b} and \eqref{eq:epp}, the desired dynamics can be given by:
\begin{align}\label{eq:b_d}
\bm{b}_d = \bm{M} (\bm{u} + \bm{G}), 
\end{align}
where $\bm{M}$ and $\bm{G}$ are defined in \eqref{eq: ss parameters}. By substituting \eqref{eq:b_d} into the QP problem \eqref{eq: BalanceControlQP}, we can obtain the optimal GRFs as the input for the low-level leg controller. The objective of the QP formulation in equation \eqref{eq: BalanceControlQP} is to find a solution that ensures the actual dynamics $\bm{A} \bm{F}$ match the desired dynamics $\bm{b}_d$. The QP-based balance controller can generally achieve the desired control input function outlined in equation \eqref{eq: PDcontrol}, thus keeping the error $\bm{e}$ within a certain range. However, if the desired dynamics vector $\bm{b}_d$ violates any of the inequality constraints, such as force limits or friction constraints, the controller may yield an optimal solution $\bm{F}^*$ that may not completely align with the desired dynamics. With this solution, the optimal dynamic ${\bm{b}_d}^*$ and $\bm{u}^*$ can be written as:
\begin{align}
&{\bm{b}_d}^* = \bm{A} \bm{F}^*, 
\\
\label{eq: u_star}
&\bm{u}^* = \bm{M}^{-1}~{\bm{b}_d}^* - \bm{G}
\end{align}
where in the appendix, we will show that the $\bm{u}^*$ remains within a bounded range.

Note that the optimal ground reaction force $\bm{F}^*$ serves as the control input for the robot, and the variable $\bm{u}^*$ acts as an input for the closed-loop dynamic. The closed-loop structure for the robot is depicted in \figref{fig: ControlDiagram_QP} (the green dashed line).

\subsection{Effects of Uncertainty on Dynamic}
If we consider uncertainty in the dynamic equation \eqref{eq: SS} and assume that the matrices $\bm{D}$ and $\bm{H}$ are not accurate, then we need to present the dynamic based on the nominal matrices $\bar{\bm{D}}$, $\bar{\bm{H}}$. The model uncertainty mostly comes from inaccurate values for mass, inertia, and foot position with respect to the center of mass. In addition, various terrains (e.g., rough terrain or soft terrain) might have different impacts on the robot, which is unknown in a practical situation. Therefore, terrain uncertainty should also be considered in the dynamic model. In this section, we solely derive our control equations based on the model uncertainty. In \secref{sec: terrain}, we will elaborate on how our proposed control system can also consider terrain uncertainty.
 
Another parameter is involved in the dynamic equation, namely the yaw angle. This angle is obtained through the state estimation, and we assumed that the state estimation has minimal uncertainty. According to the definition of matrices $\bm{D}$ and $\bm{H}$ in \eqref{eq:SS}, the inaccurate value of the dynamic parameter mentioned above reflects on the $\bm{H}$ matrix. Therefore, the dynamic equation in the presence of uncertainty can be represented as:
\begin{align} \label{eq: dynamic uncertainty}
 \bm{\dot{X}} = \bm{D}\bm{X}+ (\bar{\bm{H}}+\tilde{\bm{H}}) \bm{F} + \left[\begin{array}{c}
    \bm{0}_{6 \times 1} \\
    \bm{G} 
\end{array} \right]
\end{align}
where $\tilde{\bm{H}}$ represent the uncertainty in matrix $\bm{H}$. 
It is worth noting that according to the definition of $\bm{H}$ in equation \eqref{eq: convenient SS components}, the first six rows of $\bm{H}$ consist of zeros. Thus, we can rephrase the dynamic equation \eqref{eq: dynamic uncertainty} as follows:
\begin{align}\label{eq: uncertainty}
 \bm{\dot{X}} = \bm{D} \bm{X} + \bar{\bm{H}} \bm{F} +\bm{B} \bm{G} + \bm{B} \bm{\theta}
\end{align}
where $\bm{\theta}\in \mathbb{R}^{6}$ is the vector of uncertainty for six corresponding equations and is defined as follows:
\begin{align}
\bm{\theta} \delequal \bm{B}^T \tilde{\bm{H}}\bm{F}
\end{align} 
With reference to the state representation given by equation \eqref{eq: uncertainty}, the vector $\bm{\theta}$ can be interpreted as a time-varying disturbance affecting the body and orientation accelerations.

The uncertainty vector $\bm{\theta}$ depends on both time $t$ and $\bm{F}$. Since $\bm{F}$ is obtained through the QP problem \eqref{eq: BalanceControlQP}, it is a function of $\bm{b}_d$. Furthermore, $\bm{b}_d$ is a function of $\bm{u}$ according to \eqref{eq:b_d}. Considering that $\bm{u}$ is determined by the PD control \eqref{eq: PDcontrol}, we can conclude that $\bm{\theta}$ is a function of both the tracking error $\bm{e}$ and time $t$. As a result, for any given time $t$, it is always possible to find $\bm{\alpha}(t)\in \mathbb{R}^{6}$ and $\bm{\beta}(t)\in \mathbb{R}^{6}$ satisfying \cite{L1_adaptive}:
\begin{align}
\bm{\theta}(\bm{e},t)=\bm{\alpha}(t)||\bm{e}||+\bm{\beta}(t).
\end{align}

\subsection{Designing Adaptive Controller for Compensating the Uncertainty}\label{sec: L1_adaptive}

By incorporating $L_1$ adaptive controller, we want to design a combined controller $\bm{u}=\bm{u}_1+\bm{u}_2$, where $\bm{u}_1$ is the control input to follow the desired trajectory for the nominal model as presented in \eqref{eq: PDcontrol} and $\bm{u}_2$ is to compensate the nonlinear model uncertainties $\bm{\theta}$. Therefore, the goal is to adjust the control signal $\bm{u}_2$ so that the real model can follow the reference model. For the reference model, we employ a similar linear model described in \eqref{eq:linear_model} which, instead of $\bm{M}$, the nominal matrix $\bar{\bm{M}}$ is being used. The proposed force-based adaptive control diagram based on a balance controller is presented in \figref{fig: ControlDiagram_QP}.

The duplicate version of equation \eqref{eq: uncertainty} for state space representation presented in \eqref{fgBarDynamics} by considering combined controller $\bm{u}=\bm{u}_1+\bm{u}_2$ is as follows:
\begin{equation}
\label{eq: EtaClosedLoopUncertainty}
\dot{\bm{e}}=\bm{D}_l \bm{e}+\bm{B}\bm{u}_1 + \bm{B} (\bm{u}_2+\bm{\theta}).
\end{equation} 
Note that the vector of uncertainty $\bm{\theta}$ in equations \eqref{eq: uncertainty} and \eqref{eq: EtaClosedLoopUncertainty} are not the same since the state vector of equation \eqref{eq: uncertainty} is $\bm{X}$. In contrast, the state vector of equation \eqref{eq: EtaClosedLoopUncertainty} is system error $\bm{e}$.

The state representation for the reference model can be expressed as follows:
\begin{align}
\label{eq: ref_model}
\dot{\hat{\bm{e}}}=\bm{D}_l \hat{\bm{e}}+\bm{B} \hat{\bm{u}}_{1}+\bm{B} (\bm{u}_2+\hat{\bm{\theta}}),
\end{align}
where,
\begin{align}
\hat{\bm{\theta}}=\hat{\bm{\alpha}}||\bm{e}||+\hat{\bm{\beta}},
\end{align}
and $\hat{\bm{u}}_1$ is defined as:
\begin{equation}
\hat{\bm{u}}_1 = \begin{bmatrix}-\bm{K}_P & -\bm{K}_D\end{bmatrix} \hat{\bm{e}}. \end{equation}
To compensate the estimated uncertainty $\hat{\bm{\theta}}$, we can just simply choose $\bm{u}_2=-\hat{\bm{\theta}}$ to obtain
\begin{equation}
\dot{\hat{\bm{e}}}=\bm{D}_l \hat{\bm{e}}+\bm{B} \hat{\bm{u}}_{1}.
\end{equation}
However, $\hat{\bm{\theta}}$ typically has high frequency due to fast estimation in the adaptation law. Therefore, we employ a low-pass filter to obtain smooth control signals as follows:
\begin{align}
\label{eq: controller mu2}
\bm{u}_2=-C(s)\hat{\bm{\theta}},
\end{align}
where $C(s)$ is a second-order low-pass filter with a magnitude of 1: 
\begin{align}
\label{eq: Cs}
C(s) = \frac{{\omega_n}^2}{s^2 + 2 \zeta \omega_n s+ {\omega_n}^2} . 
\end{align}

According to \eqref{eq:b_d}, the $\bm{b}_d$ for the real model in the presence of uncertainty get the following form:
\begin{align}
\bm{b}_d = \bar{\bm{M}} (\bm{u}_1 + \bm{u}_2 + \bm{G}).
\end{align}
Respectively, $\hat{\bm{b}}_d$ for reference model is as follows:
\begin{align}
\hat{\bm{b}}_d = \bar{\bm{M}} (\hat{\bm{u}}_1 + \bm{u}_2 + \hat{\bm{\theta}} + \bm{G}).
\end{align}
The QP solver outlined in equation \eqref{eq: BalanceControlQP} allows us to obtain the optimal GRFs for the real model. Similarly, the optimal GRFs $\hat{\bm{F}}$ for the reference model can be obtained as follows:
\begin{align}
\nonumber
\hat{\bm{F}}^* =  \underset{\hat{\bm{F}} \in \mathbb{R}^{12}}{\operatorname{argmin}}   \:\:  &(\hat{\bm{A}} \hat{\bm{F}} - \hat{\bm{b}}_d)^T \bm{S} (\hat{\bm{A}} \hat{\bm{F}} - \hat{\bm{b}}_d)  \\ \label{eq: BalanceControlQP_RF} & + \gamma_1 \| \hat{\bm{F}} \|^2 + \gamma_2 \| \hat{\bm{F}} - \hat{\bm{F}}_{\textrm{prev}}^* \|^2\\
\mbox{s.t. }& \quad \:\:  \underline{\bm{d}} \leq \bm{C} \hat{\bm{F}} \leq \bar{\bm{d}} \nonumber\\
& \quad \:\: \hat{\bm{F}}_{swing}^z=0 \nonumber.
\end{align}

Define the difference between the real model and the reference model $\tilde{\bm{e}}=\hat{\bm{e}}-\bm{e}$, we then have,
\begin{align}
\dot{\tilde{\bm{e}}}=\bm{D}_l \tilde{\bm{e}}+\bm{B} \tilde{\bm{u}}_{1}+\bm{B} (\tilde{\bm{\alpha}}||\bm{e}||+\tilde{\bm{\beta}}),
\end{align}
where
\begin{align}
\tilde{\bm{u}}_{1}=\hat{\bm{u}}_{1}-\bm{u}_1,~
\tilde{\bm{\alpha}}=\hat{\bm{\alpha}}-\bm{\alpha},~
\tilde{\bm{\beta}}=\hat{\bm{\beta}}-\bm{\beta}.
\end{align}
As a result, we will estimate $\bm{\theta}$ indirectly through $\bm{\alpha}$ and $\bm{\beta}$, or the values of $\hat{\bm{\alpha}}$ and $\hat{\bm{\beta}}$ computed by the following adaptation laws based on the projection operators \cite{Lavretsky2011},
\begin{align}
\label{eq: adap_law}
\dot{\hat{\bm{\alpha}}}=\bm{\Gamma}\text{Proj}(\hat{\bm{\alpha}},\bm{y}_{\alpha}),~
\dot{\hat{\bm{\beta}}}=\bm{\Gamma}\text{Proj}(\hat{\bm{\beta}},\bm{y}_{\beta})
\end{align}
where $\bm{\Gamma} \in \mathbb{R}^{6 \times 6}$ is a symmetric positive definite matrix. The projection functions $\bm{y}_{\alpha}\in \mathbb{R}^{6}$ and $\bm{y}_{\beta}\in \mathbb{R}^{6}$ are:
\begin{align}\label{eq:proj_fun}
\bm{y}_{\alpha}&=-{\bm{B}}^T \bm{P}\tilde{\bm{e}}||\bm{e}||, \nonumber \\
\bm{y}_{\beta}&=-{\bm{B}}^T \bm{P}\tilde{\bm{e}},
\end{align}
where $\bm{P}\in \mathbb{R}^{12 \times 12}$ is a positive definite matrix that is defined according to the stability criteria using the Lyapunov equation. Moreover, the system's stability proof is provided in the appendix. 
\section{Adaptive Force-based Control using MPC} \label{sec: adaptive MPC}

Model predictive control (MPC) has been widely used across various fields, from finance to robotics. One of MPC's main advantages is its ability to handle complex systems with multiple inputs and outputs while considering hard control constraints \cite{Mayne2014}. 
MPC has also been applied to quadruped robots, providing stable locomotion \cite{DiCarlo2018}. Thanks to dynamic prediction in MPC, using the same control framework can achieve different dynamic locomotion gaits.
However, MPC's limitations become evident when dealing with significant uncertainty in the dynamic model. For instance, in the case of a quadruped robot carrying an unknown heavy load, MPC fails to track the desired state trajectory, resulting in unstable behavior and deviation from the desired trajectory, especially with dynamic gaits like bounding. 
Furthermore, the ability of a robot to traverse soft terrain where the impact model is unknown can present a significant challenge. Our proposed approach can tackle this challenge effectively, and we will discuss how it handles the terrain unknown impact model in \secref{sec: terrain}.

In the previous section \secref{sec: adaptive control}, we presented an adaptive force-based control framework based on the balance controller. The balance controller relies on a quadratic program (QP) solver, which is simple to put into practice and well-suited for slow and safe motions like standing and quasi-static walking. Additionally, the balance controller is an instantaneous control technique, meaning it does not predict the robot's future movement. As a result, the balance controller proves to be ineffective in fast-paced, highly dynamic scenarios. On the other hand, MPC has shown great potential in handling agile motions, even when it comes to underactuated gaits such as bounding.

In this section, we will present a novel control architecture to integrate adaptive control into the MPC framework. By this proposed framework, we can achieve fast and robust locomotion in the presence of uncertainties. This framework can also be extended to accommodate various dynamic gaits in legged robots, such as trotting and bounding. As discussed in a previous section, our approach is not restricted to a specific type of adaptive control. Still, we have chosen to utilize $L_1$ adaptive control, which has demonstrated advantages over other adaptive control techniques. The first step in integrating $L_1$ adaptive control and MPC is understanding the importance of a reference model and the challenges in synchronizing the real and reference models. We then present our proposed adaptive MPC, which combines conventional MPC \cite{DiCarlo2018} with adaptive control. Finally, we address the challenge of real-time computation while having two MPCs in our control system. We will elaborate on how to adjust the frequency of each control component in an optimized manner to allocate enough computation resources for critical control parts and achieve real-time computation.

\subsection{Reference Model}
Our method aims to design a combined controller based on MPC and $L_1$ adaptive control that the real model follows the reference model. In accordance with our previous discussion in \secref{sec: L1_adaptive}, the combined controller incorporates a control signal $\bm{u}_2$ to account for model uncertainty, as indicated in equation \eqref{eq: EtaClosedLoopUncertainty}. In this section, the auxiliary control signal for this purpose is $\bm{u}_a \in \mathbb{R}^{6}$; thus, the uncertain dynamic equation \eqref{eq: uncertainty} can be rewritten as follow:
\begin{align}\label{eq: model_with_aux_signal}
 \bm{\dot{X}} = \bm{D} \bm{X} + \bar{\bm{H}} \bm{F} + \bm{B} \bm{G} + \bm{B} (\bm{u}_a + \bm{\theta}).
\end{align}
The reference model is similar to the quasi-linear model described in \eqref{eq: SS} which, instead of $\bm{H}$, the nominal matrix $\bar{\bm{H}}$ is being used. The proposed adaptive MPC diagram is presented in \figref{fig: ControlDiagram_MPC}.

\begin{figure}[t!]
	\center	\includegraphics[width=1.0\linewidth]{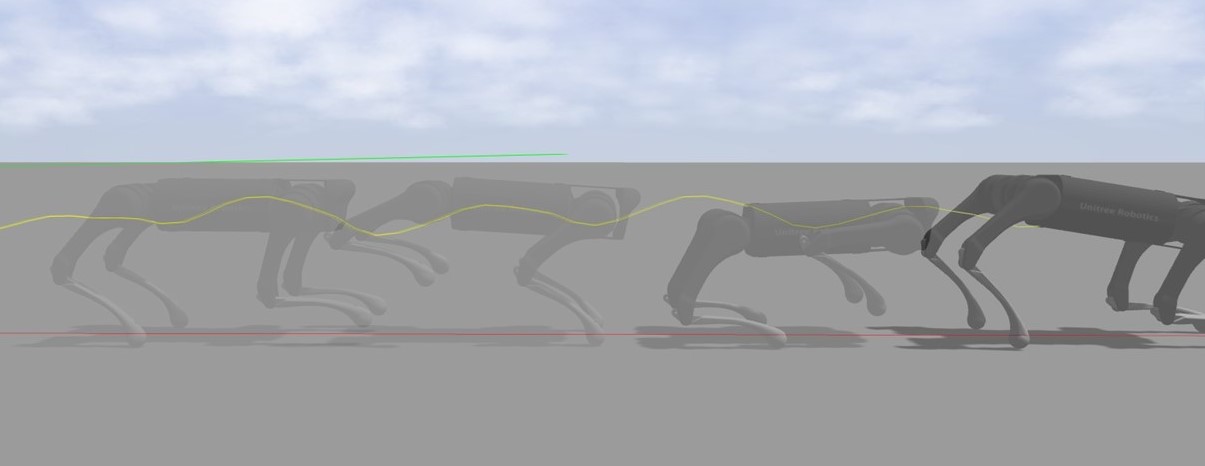}
	\caption{\textbf{Motion snapshot of the robot with bounding gaits.} A simple controller cannot easily predict The quadruped's center of mass motion (yellow line). This illustration can represent the importance of using MPC for reference model}
	\label{fig: bounding snapshot}
\end{figure}

We consider a reference model for $L_1$ adaptive control that arises from MPC. The MPC method is computationally expensive, but replacing it with other simpler control methods, such as the balance controller, while simulating the robot's performance using dynamic gaits such as bounding is impossible. The reason is that in bounding gait, the robot's two feet on either the front or rear side touch the ground at each time step, making it challenging to accurately control the height and pitch angle. The MPC approach balances the error in the height and pitch angle and, based on the predicted dynamics of the system in the future, computes the optimal ground reaction forces. As seen in \figref{fig: bounding snapshot}, the center of mass (COM) height oscillates around the desired value. Thus, the underactuated nature of certain gaits, such as bounding, necessitates the use of MPC as the control system for the reference model.

When implementing MPC for a reference model, one challenge is ensuring that the reference model is synchronized with the real model. This is particularly important when the robot performs a gait with a periodic behavior, such as bounding (see \figref{fig: bounding snapshot}). In order to correctly compare the real model with the reference model, both should have the same gait schedule. Additionally, the adaptive MPC proposed for legs in the stance phase is independent of the swing leg control. However, the foot position is crucial in calculating the moment of ground reaction force around the center of mass. Therefore, to maintain consistency between the real and reference models, it is important to ensure that the real robot's foot position is fed into the reference model as shown in \figref{fig: ControlDiagram_MPC}.

The reference model can be expressed as follows:
\begin{align}
\label{ref_model_mpc}
\dot{\hat{\bm{X}}} = \bm{D}\hat{\bm{X}} + \bm{\bar{H}} \hat{\bm{F}} + \bm{B} \bm{G} + \bm{B}(\bm{u}_a+\hat{\bm{\theta}}),
\end{align}
where
\begin{align}
\hat{\bm{\theta}}=\hat{\bm{\alpha}}||\bm{e}||+\hat{\bm{\beta}}.
\end{align}
In this case, similar to  \secref{sec: adaptive control}, we use a second-order low-pass filter, same as \eqref{eq: Cs}. Therefore, the auxiliary control signal would be:
\begin{align}
\label{eq: mu2_mpc}
\bm{u}_a=-C(s)\hat{\bm{\theta}}.
\end{align}

By defining the difference between the real model and the reference model $\tilde{\bm{X}}=\hat{\bm{X}}-\bm{X}$, we then have:
\begin{align}\label{eq:difference}
\dot{\tilde{\bm{X}}}=\bm{D}\tilde{\bm{X}}+\bar{\bm{H}}\tilde{\bm{F}}+\bm{B}(\tilde{\bm{\alpha}}||\bm{e}||+\tilde{\bm{\beta}}),
\end{align}
where
\begin{align}
\tilde{\bm{F}}=\hat{\bm{F}}-\bm{F},~
\tilde{\bm{\alpha}}=\hat{\bm{\alpha}}-\bm{\alpha},~
\tilde{\bm{\beta}}=\hat{\bm{\beta}}-\bm{\beta}.
\end{align}
Since the desired trajectory for both the real model and the reference model is the same ($\bm{X}_d = \hat{\bm{X}}_d$), the difference between the real model and reference model can be defined as:
\begin{align}
\tilde{\bm{X}} = (\hat{\bm{X}} - \hat{\bm{X}}_d) - (\bm{X} - \bm{X}_d) = \hat{\bm{e}} - \bm{e} = \tilde{\bm{e}}.
\end{align}
Therefore, equation \eqref{eq:difference} is equal to the following equation:
\begin{align}
\dot{\tilde{\bm{e}}}=\bm{D}\tilde{\bm{e}}+\bar{\bm{H}}\tilde{\bm{F}}+\bm{B}(\tilde{\bm{\alpha}}||\bm{e}||+\tilde{\bm{\beta}}).
\end{align}
The adaption laws and projection functions for computing the value of $\bm{\alpha}$ and $\bm{\beta}$ are the same as equations \eqref{eq: adap_law} and \eqref{eq:proj_fun}, respectively. Moreover, the stability of the control system can be proven using the same logic provided in the appendix. 

\subsection{Adaptive MPC}
After computing the auxiliary control signal $\bm{u}_a$ using the adaptive controller presented in the previous subsection, we will integrate the $\bm{u}_a$ with the conventional MPC for legged locomotion \cite{DiCarlo2018} and propose our adaptive MPC framework. We treat the auxiliary control signal $\bm{u}_a$ as a residual vector in the system's equation to compensate for dynamic uncertainty. Therefore, the $\bm{u}_a$ should be combined into the state vector and the equation \eqref{eq: model_with_aux_signal} can be written as follow:
\begin{align}
\label{eq: combined SS}
\bm{\dot{\eta}} = \bm{D}^{e} \bm{\eta} + \bar{\bm{H}}^{e} \bm{F} +  \bm{B}^{e} \bm{\theta} 
\end{align}
with the following extended matrices:
\begin{align}
\label{eq:components}
&\bm{\eta} = \left[\begin{array}{c} 
    \bm{X}^c \\ 
    \hline
    \bm{u}_a
    \end{array} \right] \in \mathbb{R}^{19}
\\ \nonumber
&\bm{D}^{e} = \left[\begin{array}{@{}c|c@{}} 
        \begin{matrix}
           \bm{D}^{c}_{13 \times 13}
        \end{matrix}
        & \begin{matrix}
            \bm{0}_{6 \times 6} \\
            \bm{1}_{6 \times 6} \\
            \bm{0}_{1 \times 6} \\
        \end{matrix}
    \\
    \hline
    \bm{0}_{6 \times 13} & \bm{0}_{6 \times 6}
\end{array} \right] \in \mathbb{R}^{19 \times 19}
\\ \nonumber
&\bar{\bm{H}}^{e} = \left[\begin{array}{c}
    \bar{\bm{H}}^c \\
    \hline
    \bm{0}_{6 \times 12}
\end{array} \right] \in \mathbb{R}^{19 \times 12}
\\ \nonumber
&\bm{B}^{e} = \left[\begin{array}{c}
    \bm{B} \\
    \hline
    \bm{0}_{7 \times 6}
\end{array} \right] \in \mathbb{R}^{19 \times 6}
\end{align}
where $\bar{\bm{H}}^{c}$ is the nominal value of $\bm{H}^{c}$. The definition of $\bm{X}^{c}$, $\bm{D}^{c}$, and $\bm{H}^{c}$ can be found in \eqref{eq: convenient SS components}.  Although $\bm{u}_a$ is considered a part of the state vector in \eqref{eq: combined SS}, it is just a residual vector for compensating dynamic uncertainty. Therefore, $\bm{u}_a$ is constant in the state space equation and over the horizons. To this end, the components associated with $\bm{u}_a$ in matrices $\bm{D}^{e}$ and $\bar{\bm{H}}^{e}$ are assigned zero, which means $\dot{\bm{u}}_a = 0$. Note that the value of $\bm{u}_a$ will be updated according to the adaptive law, but it is constant during the prediction horizons.

The state representation in \eqref{eq: combined SS} is also convenient for discretization methods such as zero-order hold \cite{Fadali2012} for MPC.
Therefore, our adaptive MPC can be designed according to \eqref{eq:mpc} and based on the following discrete-time dynamic:
\begin{align}\label{eq: combined_discrete_dynamic}
 \bm{\eta}_{i+1} = {\bm{D}^{e}}_{t,i} \bm{\eta}_{t,i} + {\bar{\bm{H}}^{e}}_{t,i} \bm{F}_i  
\end{align}

\subsection{Real-time Computation} \label{sec: real-time computation}
The main challenge in executing our proposed adaptive MPC framework is ensuring that the computation required is fast enough for hardware experiments in real time. If the controller is unable to perform updates at a high frequency, it could result in the robot collapsing during dynamic motion. The control system comprises two MPCs, each with 13 to 19 states predicted over ten horizons. To ensure the robot's balance and allocate sufficient computation resources to each control component, we have devised a scheme, as depicted in \figref{fig: ControlDiagram_MPC}, to update each control component in an optimized manner.

The robot's sensory data updates in real-time with a frequency of $1$  $kHz$. Thus, the reference model should update with the same frequency to compare the reference model states ($\hat{\bm{X}}$) and real model states ($\bm{X}$) correctly. The yellow dashed line in \figref{fig: ControlDiagram_MPC} indicates the update frequency for the reference model. We use the \textit{odeint} package from Boost software in C++ \cite{webpage} to solve the ODE problem associated with the dynamic equation for the reference model.

One of the critical components in our proposed framework is the adaptive MPC, which is responsible for computing the ground reaction force for the robot, as shown in \figref{fig: ControlDiagram_MPC}. Through our experimentation, we have determined that for robust locomotion with dynamic gaits, the optimal update frequency for the adaptive MPC should be 300 Hz. In contrast, the reference MPC, which plays a supporting role in the control system, is less sensitive and runs at a slower rate of 30 Hz. In addition, there is a two-millisecond delay between the running of the adaptive MPC and reference MPC to ensure sufficient computational resources are allocated to each component. This means the two MPC frameworks do not run simultaneously in our control system.
\section{Adaptation to Unknown Impact model} \label{sec: terrain}
The dynamic formulation presented in \secref{sec: adaptive control} and \secref{sec: adaptive MPC} considers the presence of model uncertainty in real-world situations. It is assumed that the terrain is hard enough to allow the robot to receive the desired force as ground reaction forces on its feet. However, this assumption may not hold if the robot walks on soft or elastic terrain with an unknown impact model, which may not generate the desired force needed for stable locomotion. Some previous studies have included terrain knowledge and contact models in their balancing controllers to address the soft terrain challenge, mainly using a spring-damper model to characterize the soft terrain \cite{Azad2015, Vasilopoulos2018}. Some control frameworks for adapting to soft terrain in real-time have also been developed using iterative learning \cite{Chang2017} and whole-body control \cite{Fahmi2020}, without prior knowledge about the terrain. This section demonstrates that the proposed method in sections \secref{sec: adaptive control} and \secref{sec: adaptive MPC} can also handle unknown impact models from terrain, allowing the robot to maintain stability while walking on soft terrains.

Equation \eqref{eq: torque_mapping}, representing the force-to-torque mapping, holds under the condition that the movement of each leg can be considered negligible. This assumption is reasonable for the stance leg on solid ground. However, when dealing with soft terrain, this mapping is not accurate. The dynamic equation for each leg is expressed as follows:
\begin{align} \label{eq: full mapping}
\bm{\tau}_{stance, i} = \bm{M}_i(\bm{q}_i) \Ddot{\bm{q}_i} + \bm{n}_i(\bm{q}_i, \Dot{\bm{q}_i}) -{\bm{J}(\bm{q}_i)}^{T} \bm{R}^{T}\bm{F}_{i}
\end{align}
where $\bm{M}_i(\bm{q}_i)$ is the inertia matrix, $\bm{n}_i(\bm{q}_i, \Dot{\bm{q}_i})$ is the nonlinear term, and $\Ddot{\bm{q}_i}$ is joints acceleration for the $i$-th leg. Remember that assumption 1 in \secref{sec: simplified robot dynamic} was considered for the rigid body dynamic equation of the robot. However, when addressing the dynamic equation of each leg, it is not valid to neglect the inertia associated with each leg. 
Furthermore, calculating the joint acceleration poses challenges due to considerable noise, making the complete computation of \eqref{eq: full mapping} difficult. This is where adaptive controllers prove to be beneficial.

Assume the computed force $\bm{F}$ by MPC in \eqref{eq: uncertainty} cannot be achieved perfectly due to walking on soft terrain. Therefore, equation \eqref{eq: uncertainty} can be rewritten as follow:
\begin{align}\label{eq: terrain uncertainty dynamic}
 \bm{\dot{X}} = \bm{D} \bm{X} + \bar{\bm{H}} (\bm{F}_a + \tilde{\bm{F}}_a) +  \bm{B} \bm{G} + \bm{B} \bm{\theta}
\end{align}
where $\bm{F}_a$ is the actual ground reaction force applied to the robot and $\tilde{\bm{F}}_a$ is the difference between the desired ground reaction force and actual reaction force. Therefore, we can reformulate equation \eqref{eq: terrain uncertainty dynamic} as follows:
\begin{align}\label{eq: uncertainty combined}
 \bm{\dot{X}} = \bm{D} \bm{X} + \bar{\bm{H}} \bm{F}_a + \bm{B} \bm{G} + \bm{B} (\bm{\theta} + \bm{\theta}_F).
\end{align}
where the uncertainty vector $\bm{\theta}_F$ is defined as follow:
\begin{align}
\bm{\theta}_F \delequal \bm{B}^T \bar{\bm{H}} \tilde{\bm{F}}_a 
\end{align}
The equation \eqref{eq: uncertainty combined} is in the form of equation \eqref{eq: uncertainty}, which uses actual ground reaction force instead of desired ground reaction force. Therefore, all formulations for implementing adaptive controllers are also valid for a situation with an unknown impact model.
\section{Results} \label{sec: Results}

In this section, we validate our control approach in simulation and hardware experiments on a Unitree A1 robot. All the hardware experiments' computations run on a single PC (Intel  i7-6500U, 2.5  GHz, 64-bit). For simulation, the control system is implemented using ROS Noetic in the Gazebo 11 simulator, which provides a high-fidelity simulation of the A1 robot. A video showcasing the results accompanies this paper\footnote{\url{https://youtu.be/5t1mSh0q3lk}}.

\begin{figure*}[t!]
	\centering
	\subfloat[gravel]{\includegraphics[width=0.24\linewidth]{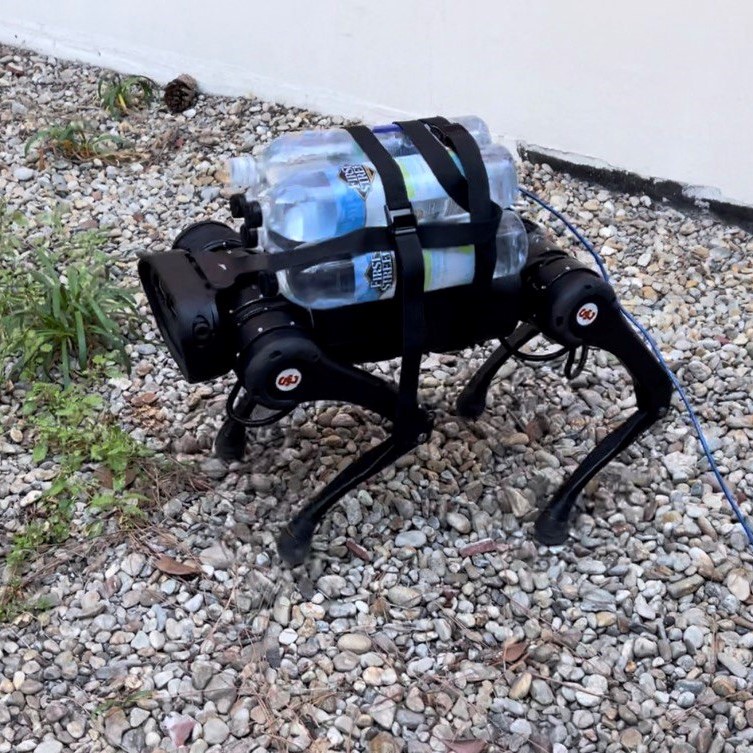}}
	\hfill
	\subfloat[grass]{\includegraphics[width=0.24\linewidth]{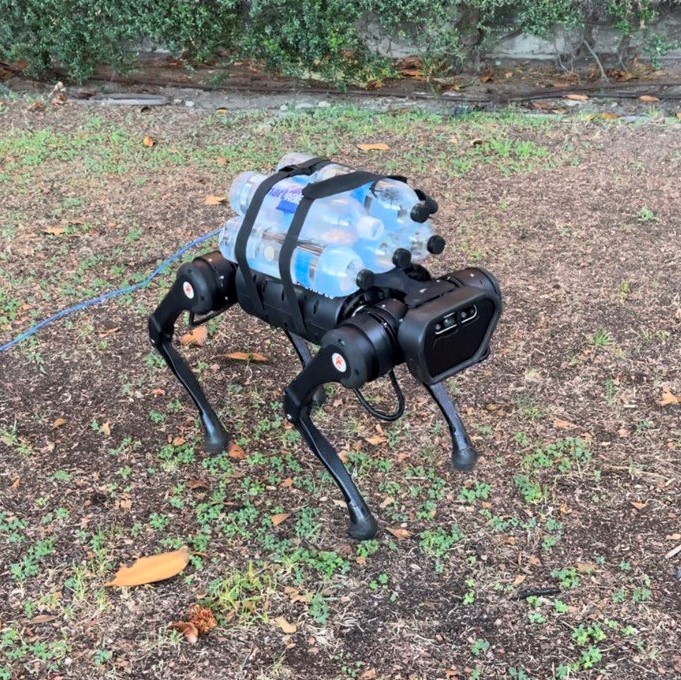}}
	\hfill
 	\subfloat[uneven terrain]{\includegraphics[width=0.24\linewidth]{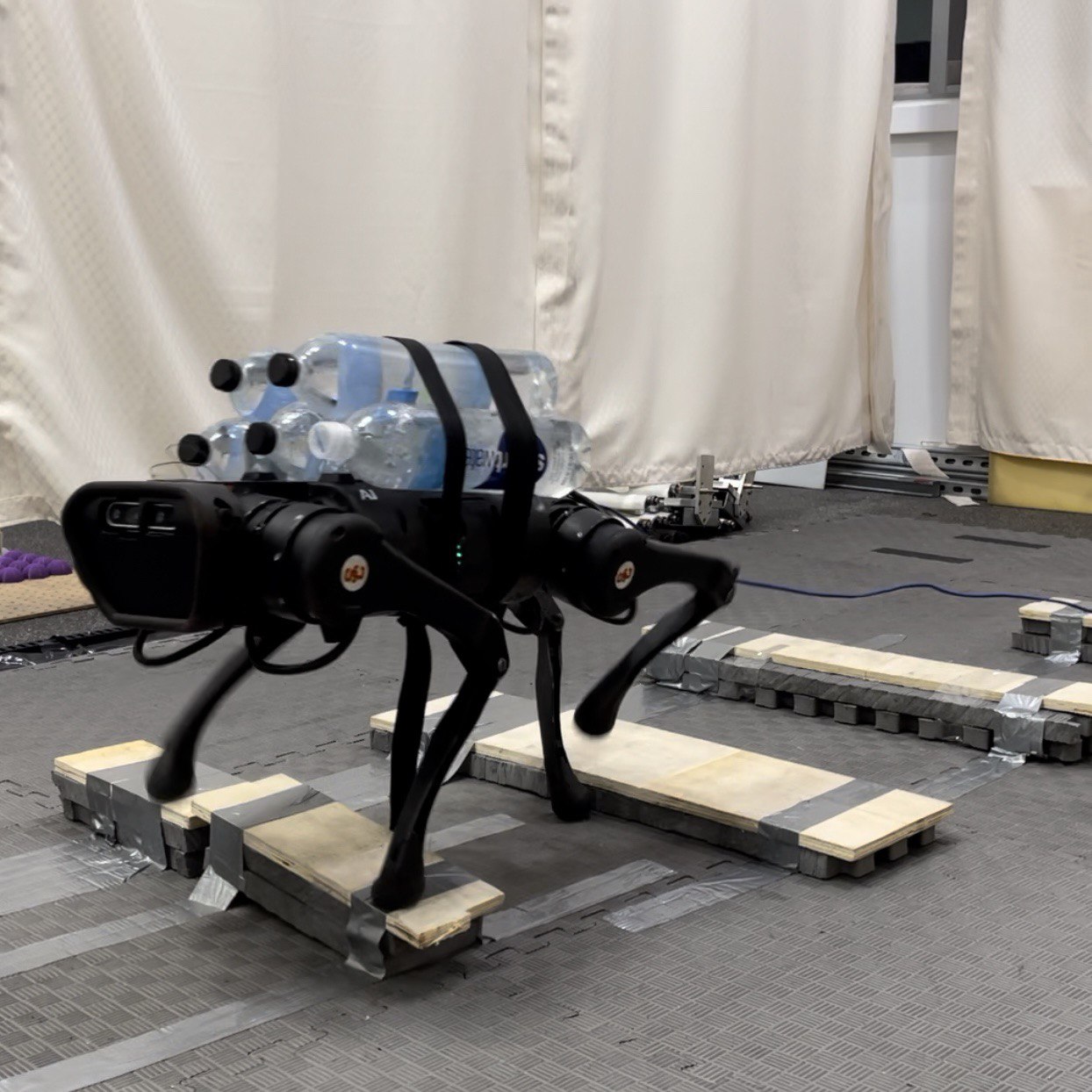}}
    \hfill
    \subfloat[high-sloped terrain]{\includegraphics[width=0.24\linewidth]{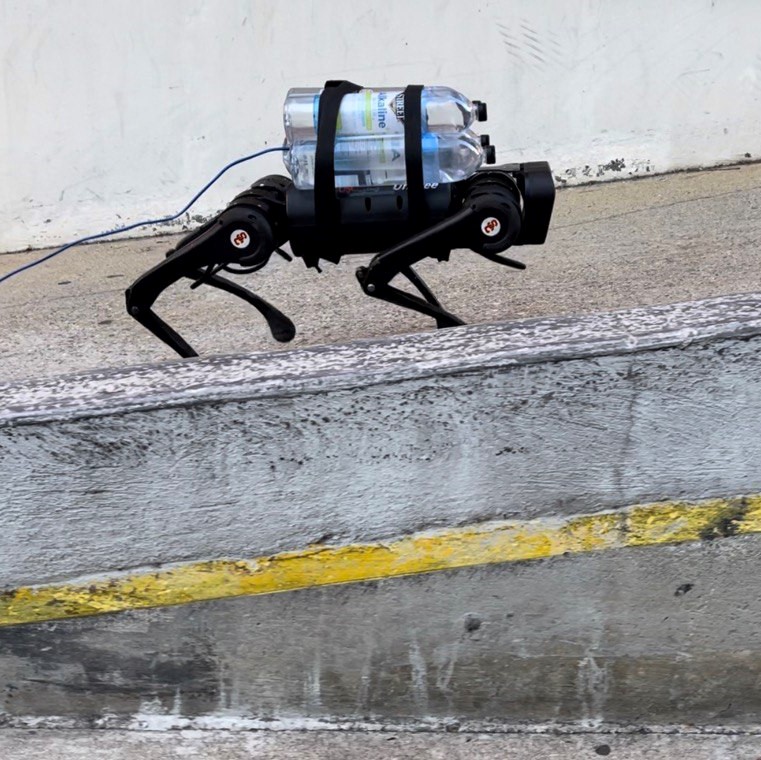}}
	
	\caption{\textbf{Navigating different terrain using our proposed adaptive MPC while carrying an unknown heavy load.} a) gravel, b) grass, c) rough terrain, d) high-sloped terrain.}
	\label{fig: terrain experiment}
		\vspace{-1em}
\end{figure*}

We set the control parameters for MPC, the adaption law, and the low-pass filter as presented in Table \ref{tab:parameter}. We use one set of parameters for all the experiments with different locomotion gaits, indicating that our approach is easily generalizable. The following subsections will introduce different experiment results in terms of model and environment uncertainty (see \figref{fig: terrain experiment}). In each experiment, the robot starts by using a balance controller to stand up and then switches to the MPC framework for walking or running.

\begin{table}[bt!]
	\centering
	\caption{Controller Setting}
	\label{tab:parameter}
	\begin{tabular}{|c|c|}
		\hline
		 Parameter & Value \\ [1ex] \hline
		
		$\bm{Q}$ & \makecell{$diag(2.5, 2.5, 20, 0.25, 0.25, 1.5, $   \\    $0.2, 0.2, 0.2, 0.1, 0.1, 0.3)$} \\ [1ex] \hline
            $\bm{R}$  & $\gamma_1 \bm{I}_{12}$ \\ [1ex] \hline 
            $\bm{\Gamma}$ & $diag(1, 1, 5, 2, 5, 1) \times         10^3$ \\ [1ex] \hline
		$\omega_n$ & 60  \\ [1ex] \hline
            $\zeta$ & 0.7 \\ [1ex] \hline
	\end{tabular}
\end{table} 

\subsection{Comparative Analysis}
In order to evaluate the performance of our proposed adaptive MPC method, we conduct a comparative experiment with the conventional MPC method presented in \cite{DiCarlo2018}. The objective is to understand the advantages of integrating the adaptive controller into MPC for quadrupedal locomotion.

\vspace{5pt}
\subsubsection{Walking with significant model uncertainty}
\begin{figure}[t!]
	\centering
	\subfloat[non-adaptive controller]{\includegraphics[width=0.48\linewidth]{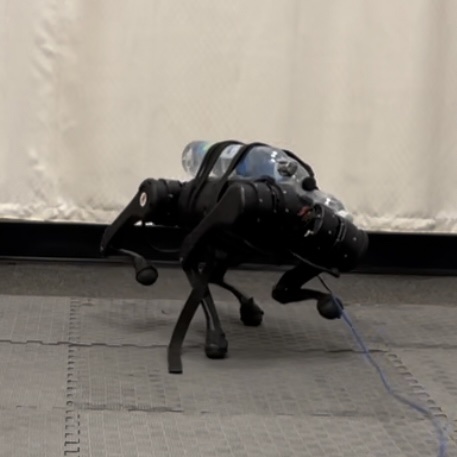}}
	\hfill
	\subfloat[adaptive controller]{\includegraphics[width=0.48\linewidth]{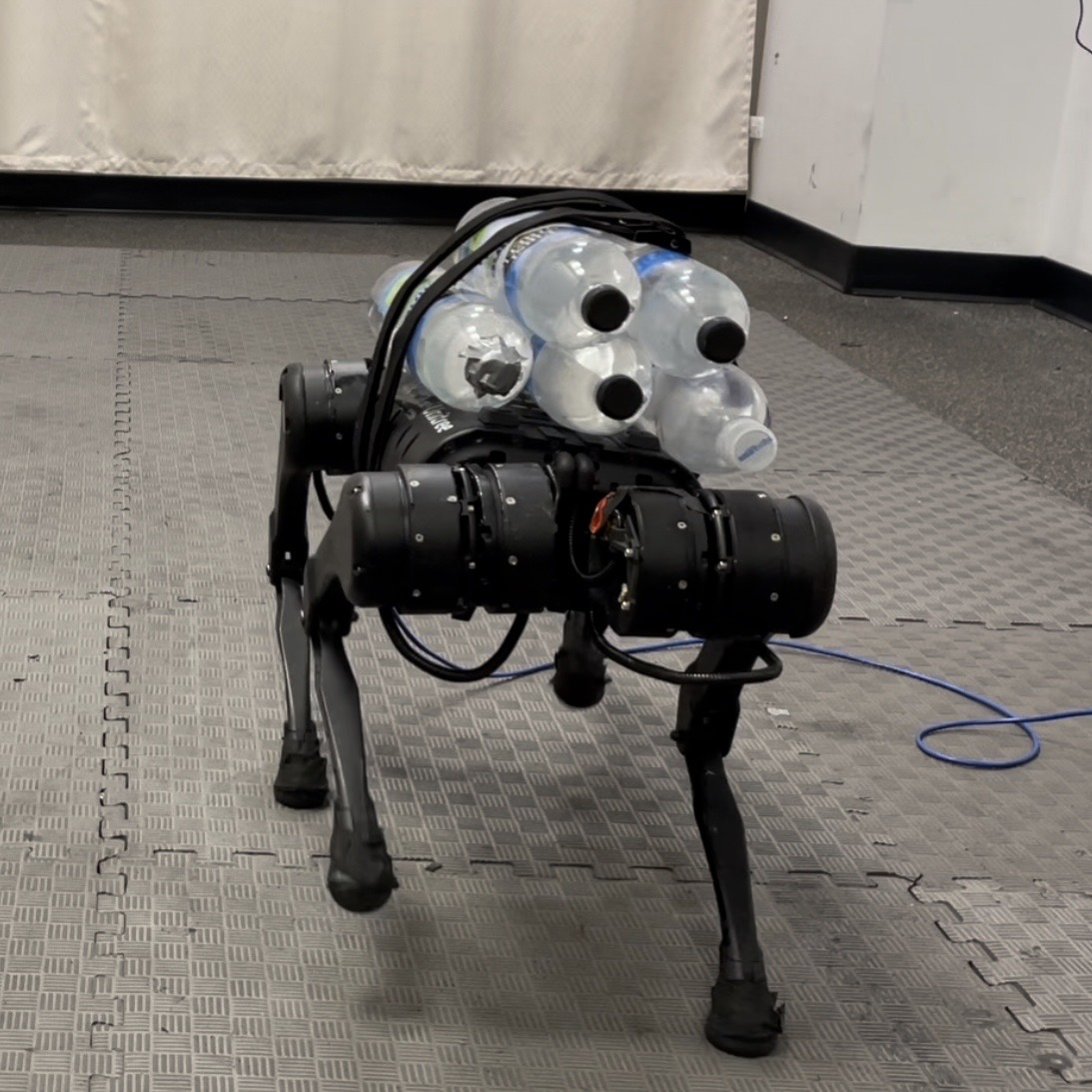}}
	\hfill
	\subfloat[]{\includegraphics[width=1\linewidth]{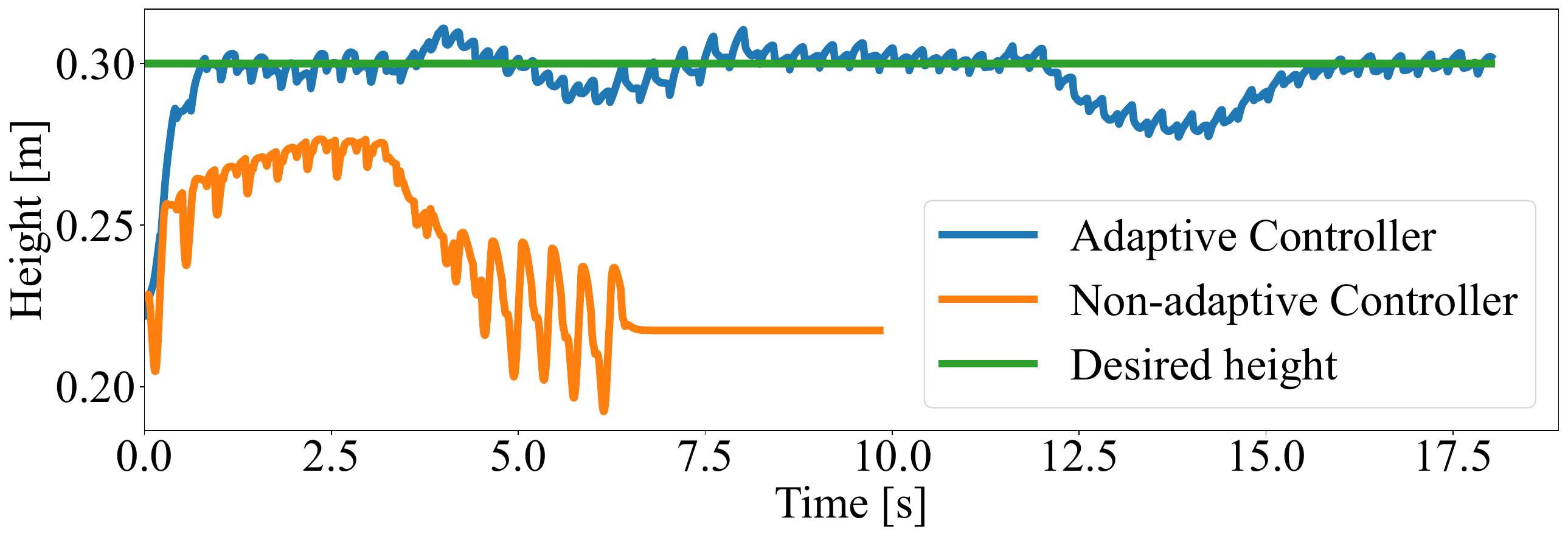}}
        \hfill
	\subfloat[]{\includegraphics[width=1\linewidth]{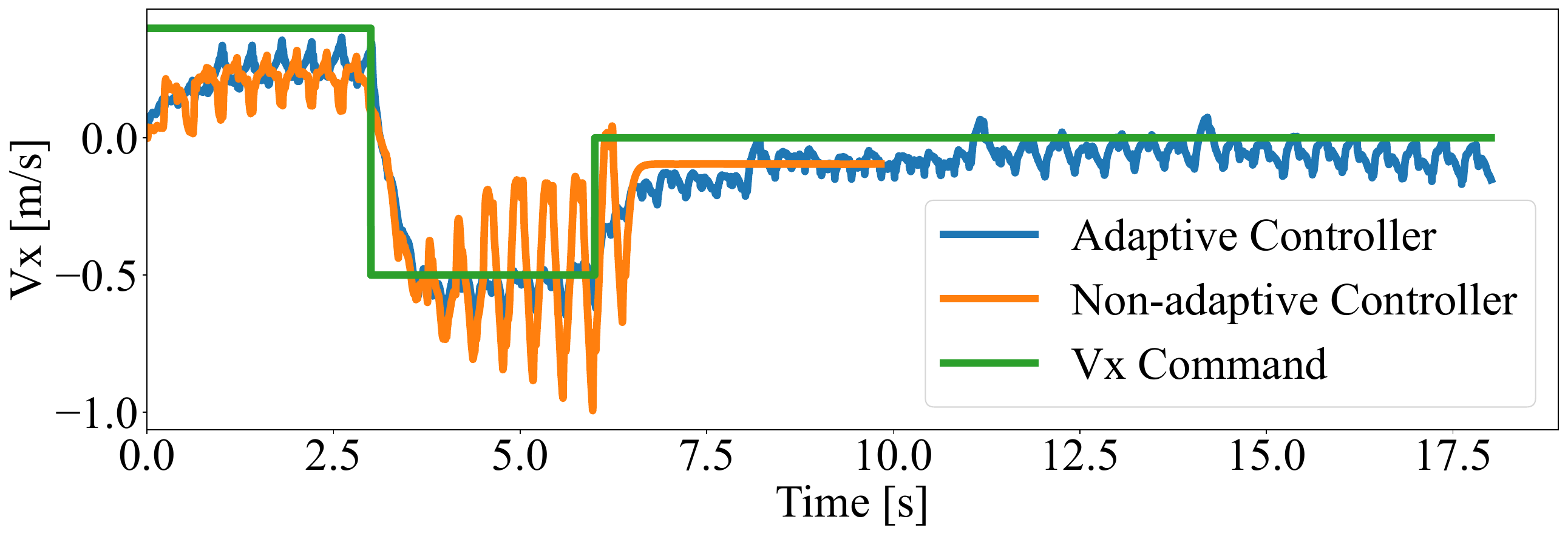}}
    \hfill
	\subfloat[]{\includegraphics[width=1\linewidth]{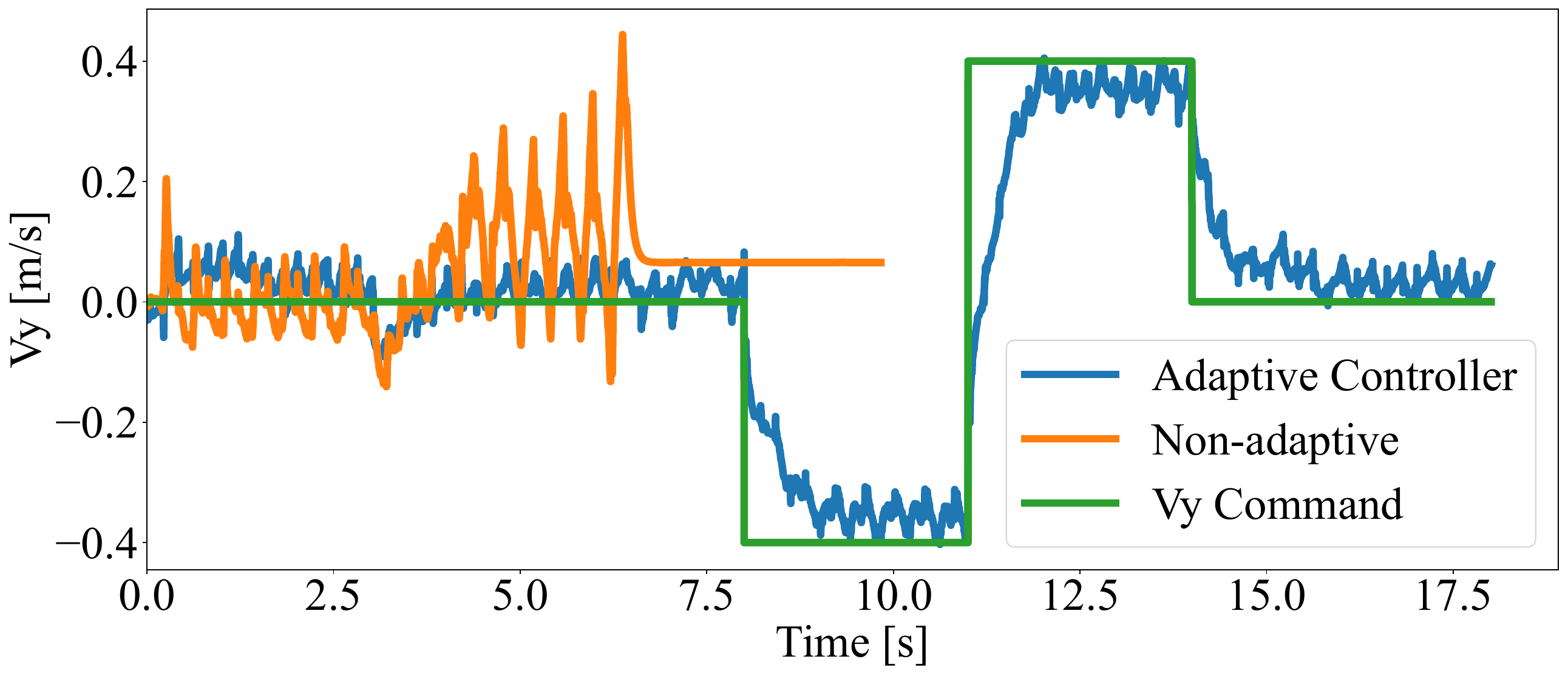}}
	
	\caption{\textbf{Comparing performance of adaptive and non-adaptive controllers.} a) Snapshots of the A1 robot with the non-adaptive controller while carrying an unknown 3 kg load and collapses, b) Snapshots of the A1 robot walking robustly with the adaptive controller while carrying an unknown 5 kg load. c) Comparative plots of the COM height, d) velocity command tracking in the x-direction, and e) y-direction for adaptive and non-adaptive controllers. The plots for the non-adaptive controller do not persist until the completion of the unit test, as the robot experiences a collapse during testing.}
	\label{fig: comparison exp}
\end{figure}

The experiment involves the robot walking and rotating in different directions, using both adaptive and non-adaptive controllers while carrying an unknown load. 
To ensure a precise comparison, we establish a unit test comprising velocity and rotation commands. This test is then executed for both adaptive and non-adaptive controllers.
The experiment results show that the adaptive controller provides robust locomotion, with excellent tracking error, even when carrying an unknown 5 kg load. On the other hand, the non-adaptive controller results in a considerable error in the COM height and eventually collapses under the weight of just a 3 kg load. The comparative results for the adaptive and non-adaptive controllers are shown in \figref{fig: comparison exp}.

\vspace{5pt}
\subsubsection{Walking on soft terrain}
\begin{figure}[t!]
	\centering
	\subfloat[non-adaptive controller]{\includegraphics[width=0.48\linewidth]{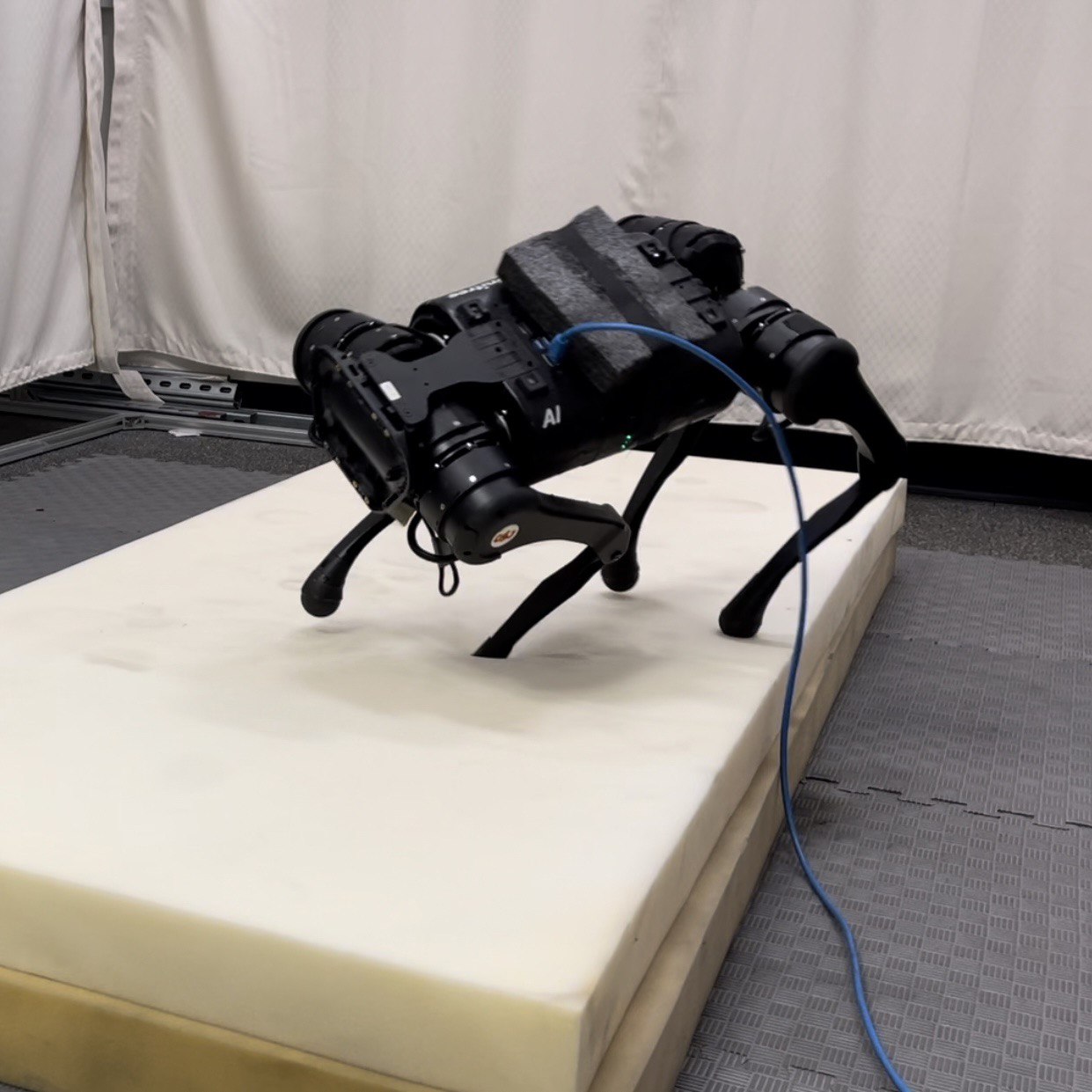}}
	\hfill
	\subfloat[adaptive controller]{\includegraphics[width=0.48\linewidth]{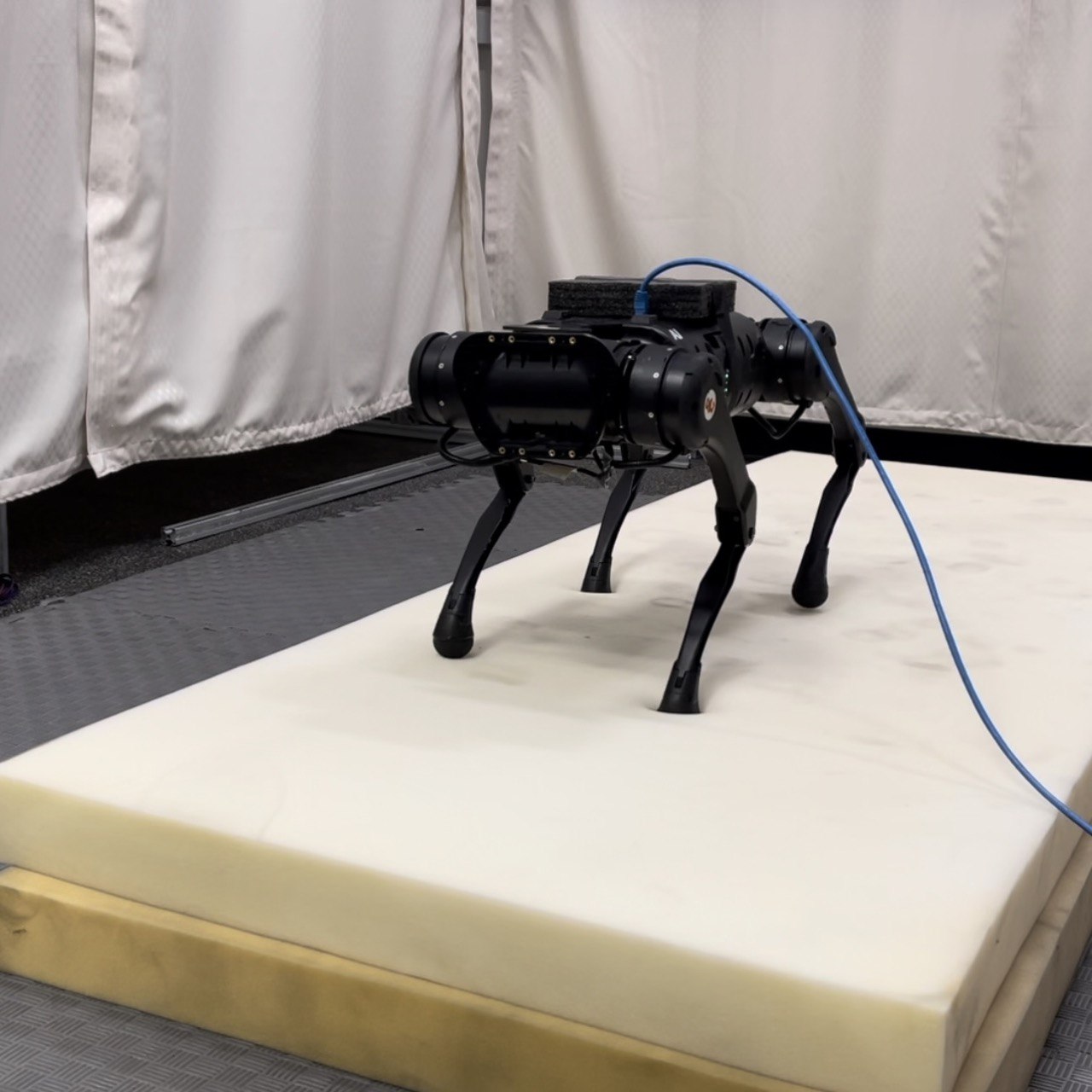}}
	\hfill
	\subfloat[]{\includegraphics[width=1\linewidth]{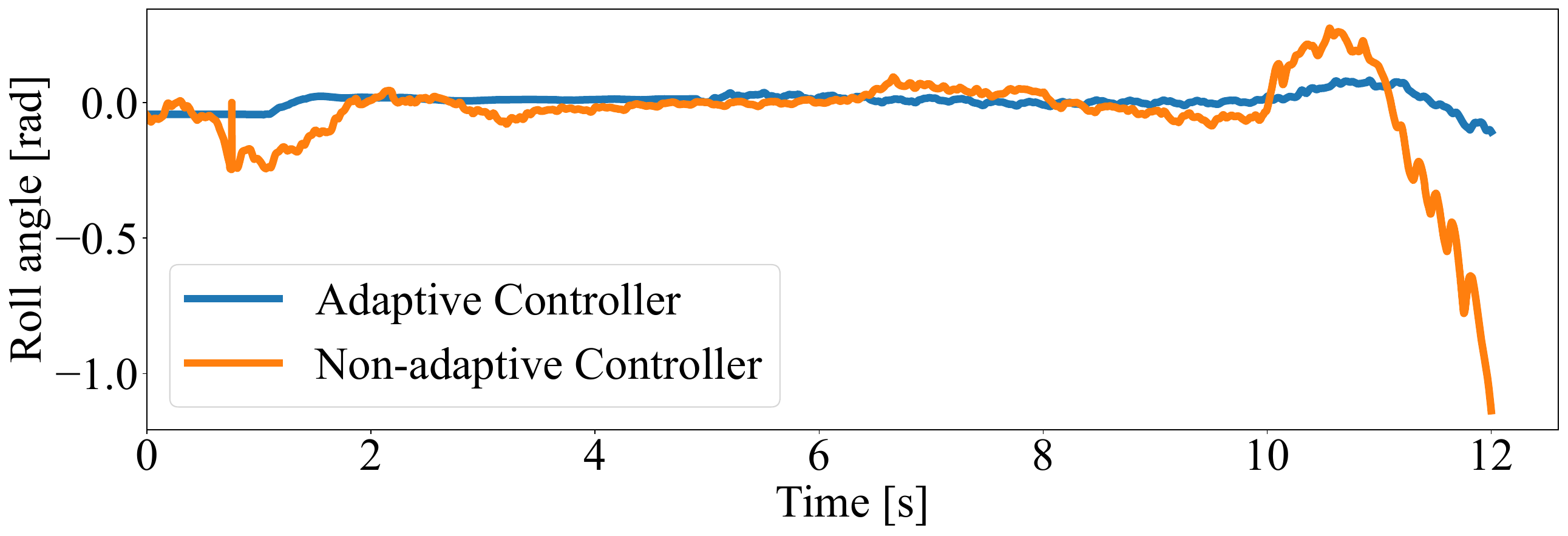}}
	\caption{\textbf{Comparing performance of adaptive and non-adaptive controllers on soft terrain.} The A1 robot tries to walk on double soft foam using a) non-adaptive and b) adaptive controllers. c) Shows the plot of the robot's roll angle.}
	\label{fig: soft terrain exp}
		\vspace{-1em}
\end{figure}

To evaluate the capability of our proposed control method in handling unknown impact models, we conducted an experiment where the robot was made for walking on a double foam, which symbolizes a soft terrain. The performance of both the adaptive and non-adaptive controllers was evaluated and compared. The results are depicted in \figref{fig: soft terrain exp}, representing the robot's roll angle. The figure clearly illustrates that the adaptive controller could maintain the robot's balance on the soft terrain. In contrast, the non-adaptive controller could not do so, leading to the collapse of the robot.

\subsection{Running with Multiple Gaits}
\begin{figure}[t!]
	\centering
	\subfloat[trotting]{\includegraphics[width=0.48\linewidth]{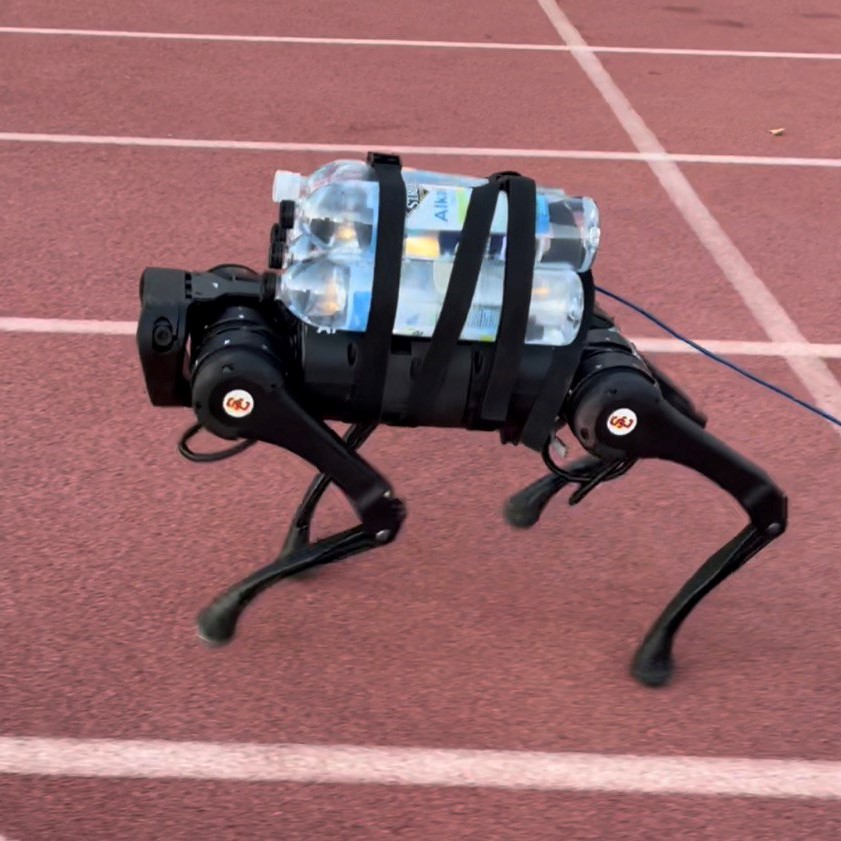}}
	\hfill
	\subfloat[bounding]{\includegraphics[width=0.48\linewidth]{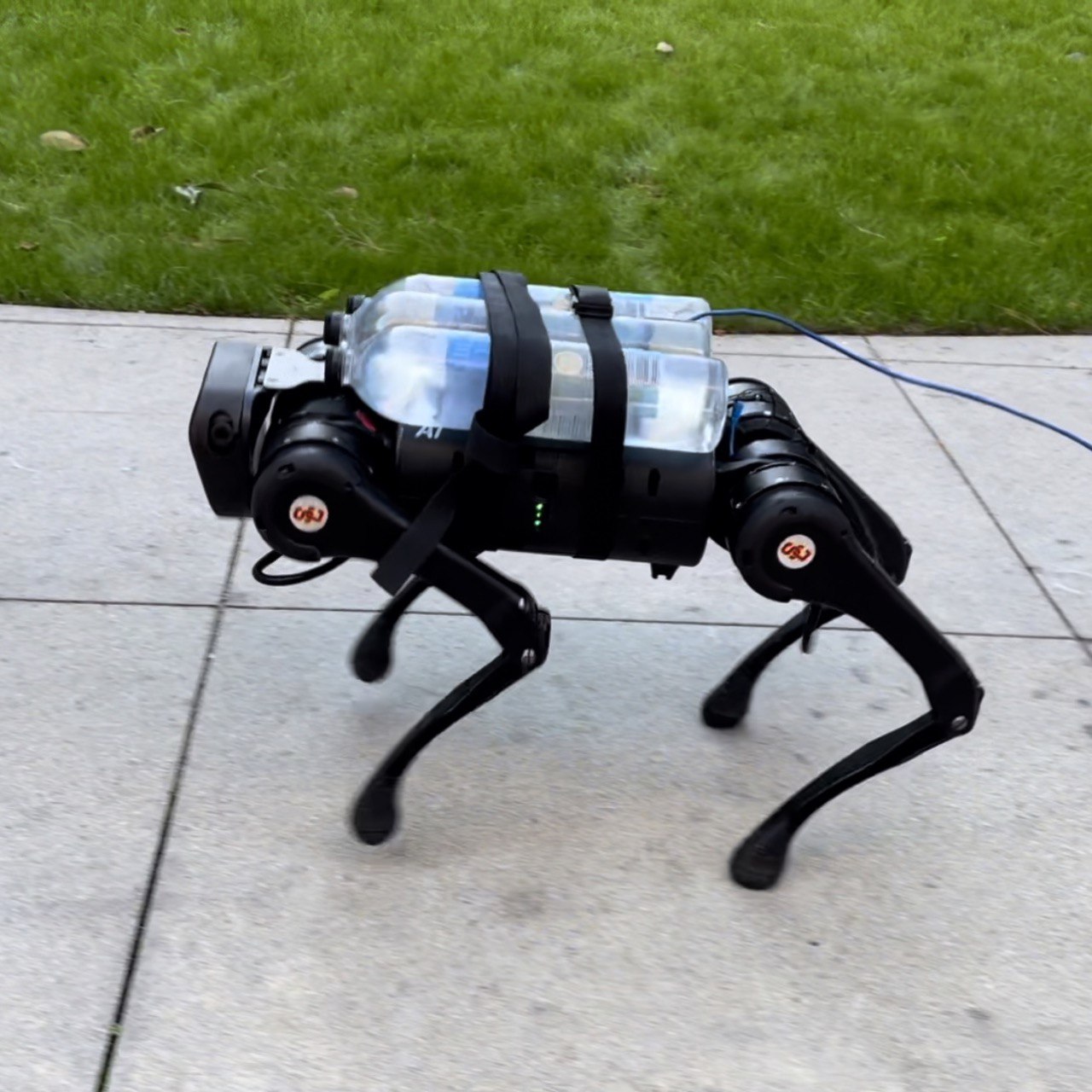}}
	\hfill
	\subfloat[]{\includegraphics[width=1\linewidth]{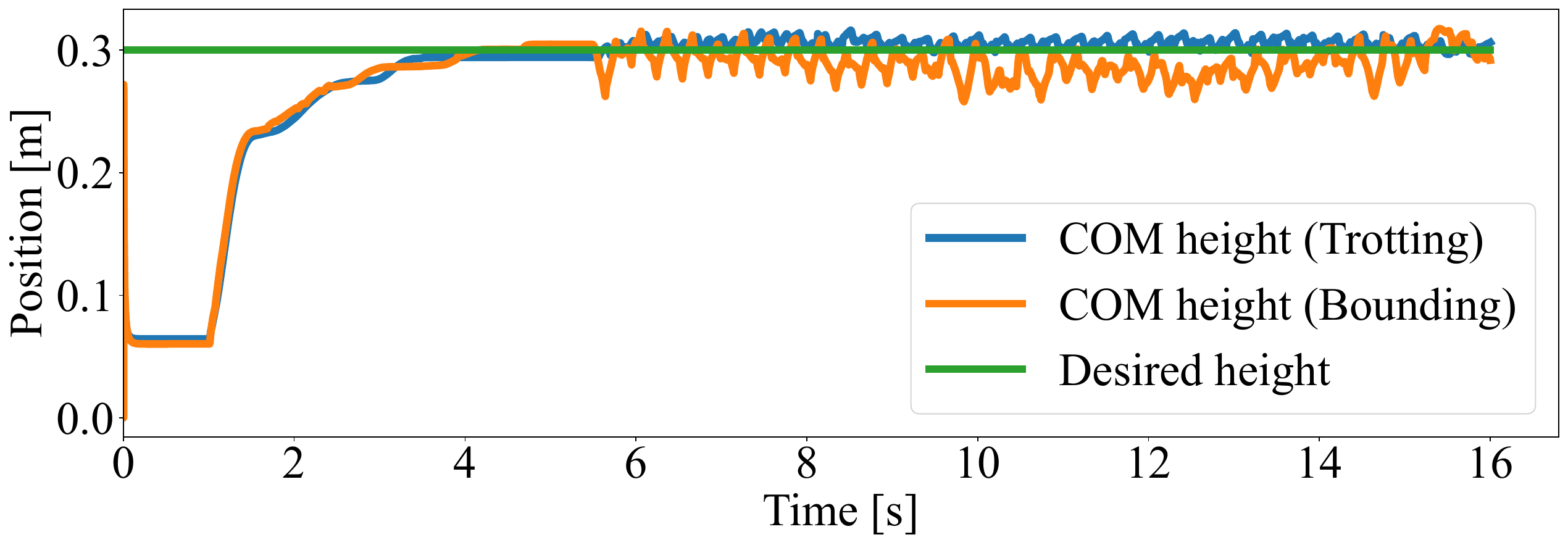}}
	\caption{\textbf{Running experiment.} The A1 robot runs with the velocity of 1 m/s using our proposed method. a) trotting gait with an unknown 5 kg load, b) bounding gait with an unknown 3 kg load, c) Plots of COM height.}
	\label{fig: running experiment}
\end{figure}

To demonstrate the superiority of our proposed approach for dynamic gaits, we conducted experiments with the robot running while carrying an unknown load. These experiments were carried out for both the trotting and bounding gaits, with an unknown load of 5 kg and 3 kg, respectively. The results of these experiments are shown in \figref{fig: running experiment}. It can be seen from the figure that the tracking of the center of mass height during the bounding gait is more unstable compared to the trotting gait, which is due to the inherent underactuated nature of the bounding gait.

\subsection{Time-varying Load}
\begin{figure}[t!]
	\centering
	\subfloat[]{\includegraphics[width=0.32\linewidth]{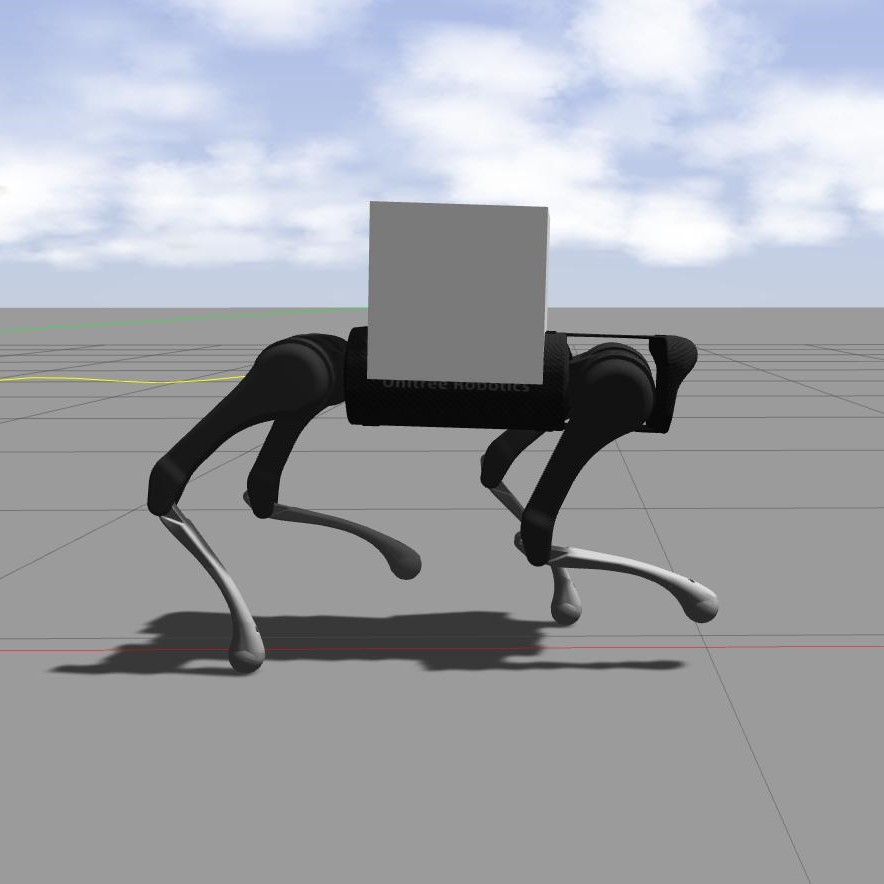}}
	\hfill
	\subfloat[]{\includegraphics[width=0.32\linewidth]{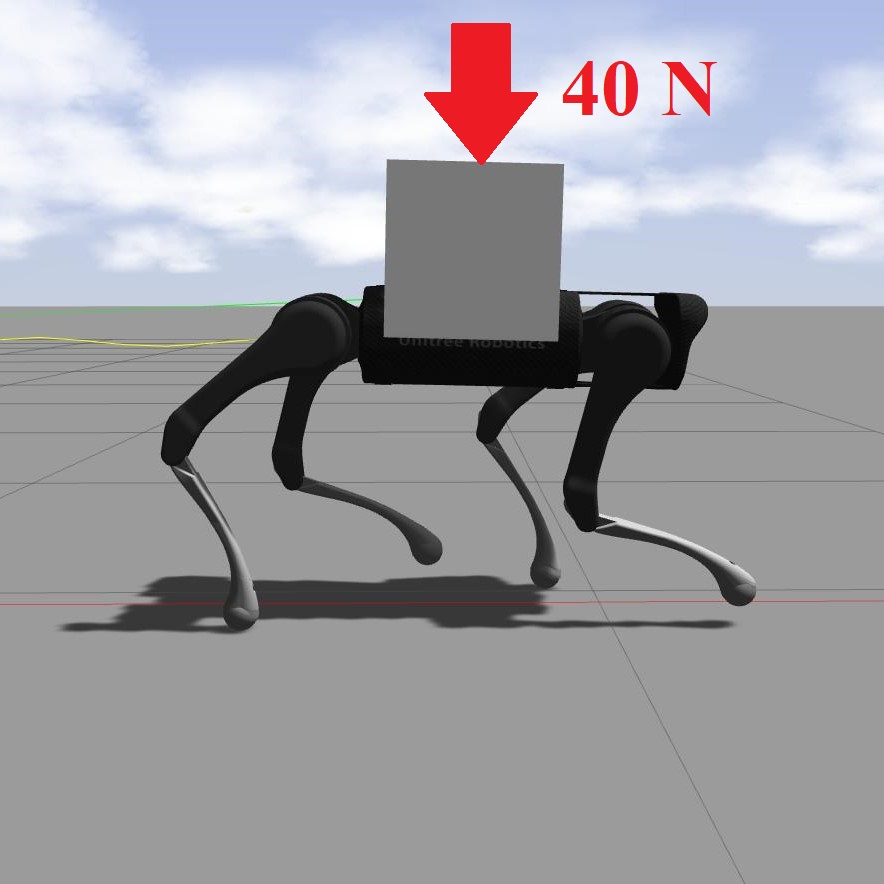}}
	\hfill
	\subfloat[]{\includegraphics[width=0.32\linewidth]{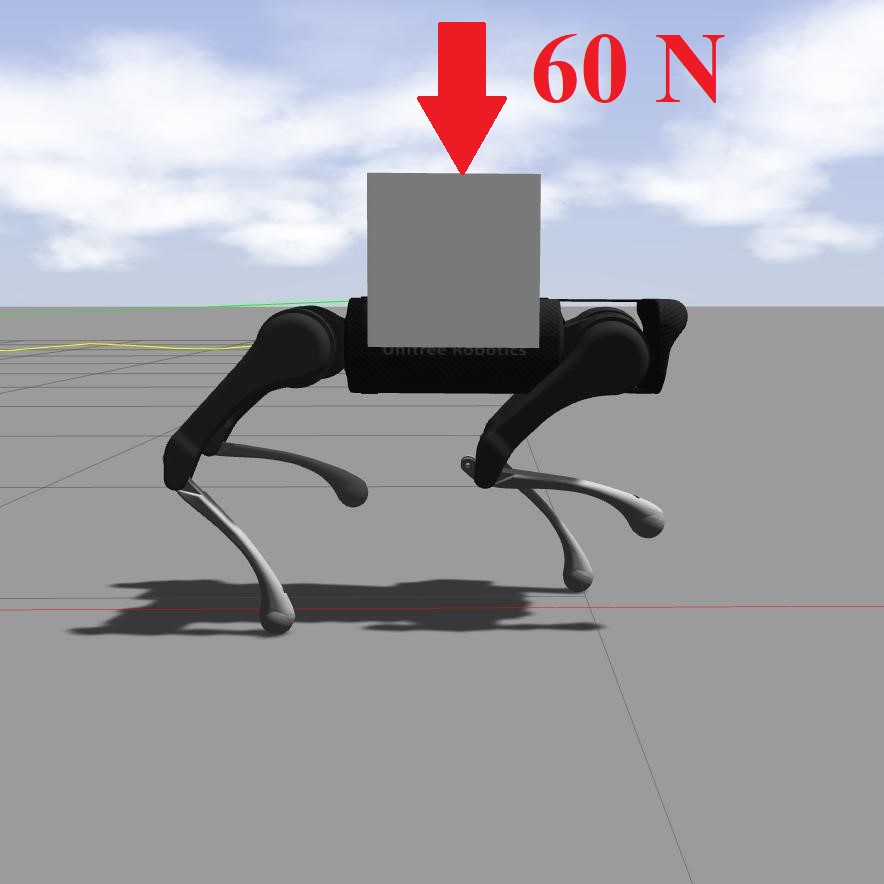}}
    \hfill
	\subfloat[]{\includegraphics[width=1\linewidth]{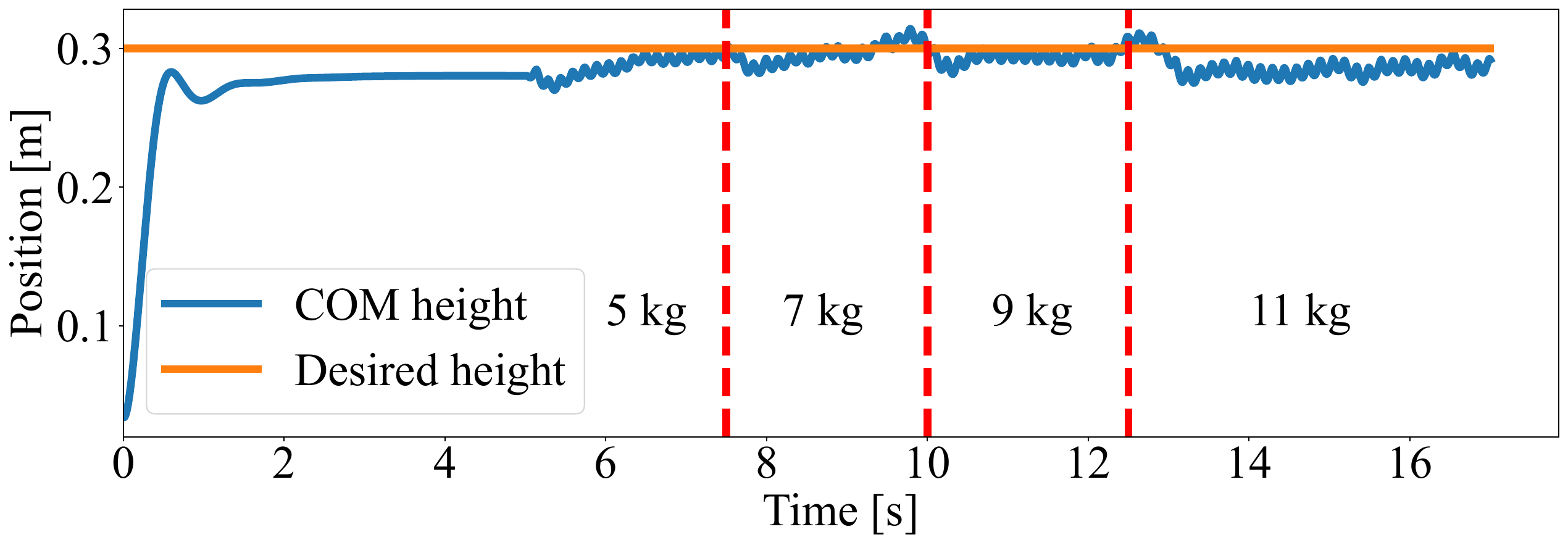}}
	\hfill
	\subfloat[]{\includegraphics[width=1\linewidth]{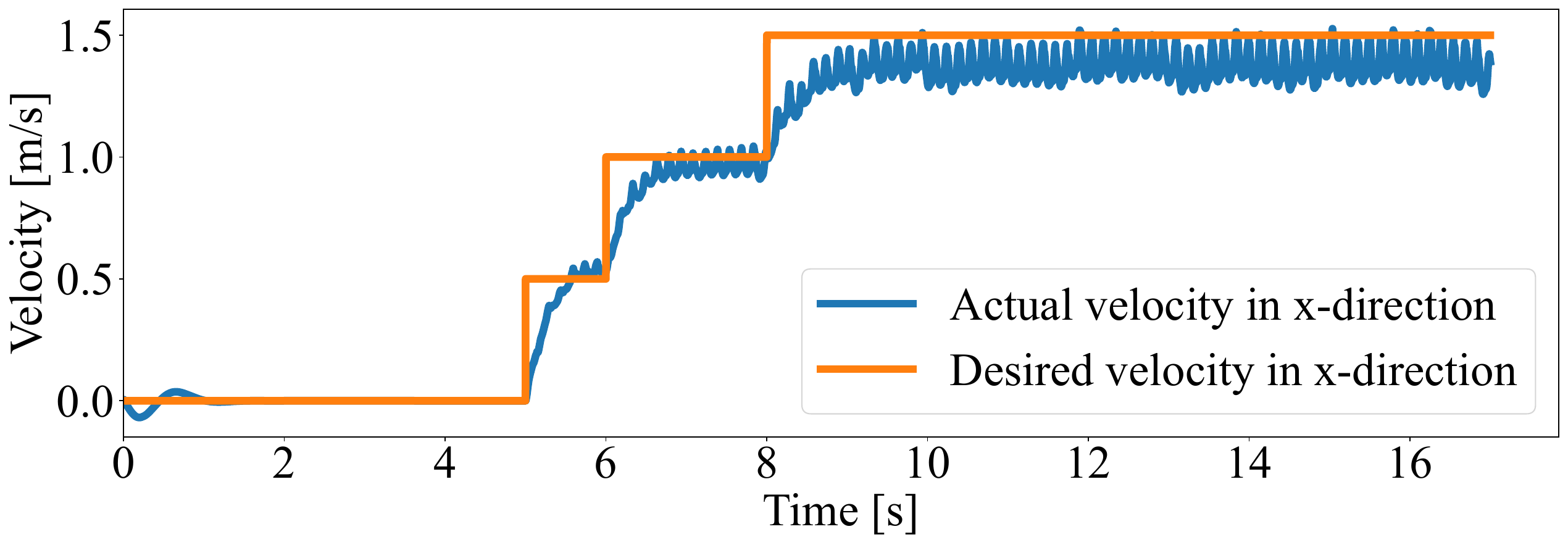}}
	
	\caption{\textbf{Simulation results for the robot carrying a time-varying load.} a) The robot starts with an unknown 5 kg load, then gradually, an unknown time-varying force will be exerted on the robot as shown in (b) and (c) while the robot's velocity increases. d) Plot of COM height, e) Robot velocity tracking in the x-direction.}
	\label{fig: time_varying result}
\end{figure}

To demonstrate the effectiveness of our proposed adaptive force control in adapting to model uncertainty, we conducted simulations where the robot carries a time-varying load of up to 92\% of its weight during walking. As shown in \figref{fig: time_varying result}, our approach can enable the robot to adapt to time-varying uncertainty. In the simulation, the robot starts with an unknown 5 kg load. While increasing the robot's velocity, the robot is subjected to a varying external force in the z-direction that rises to 60 N, resulting in an additional unknown 11 kg load. These results indicate that our proposed approach effectively handles high levels of model uncertainty.

\subsection{Terrain Uncertainty}
To demonstrate the capability of our proposed method to handle terrain uncertainty, we tested the robot navigating various terrains while carrying an unknown 5 kg load. To this end, we tried walking experiments on multiple rough terrains as well as high-sloped terrain, and we got impressive results.

\vspace{7pt}
\subsubsection{Rough terrain}
We tested the robot navigating various rough terrains such as grass and gravel. The robot walks and rotates in multiple directions while carrying an unknown 5 kg load. Some snapshots of the robot walking on diverse rough terrain are presented in \figref{fig: terrain experiment}. Our approach is based on a force controller and retains the robustness features of the baseline framework, allowing the robot to handle the rough terrain effectively.

\vspace{7pt}
\subsubsection{Sloped terrain}
To enable the robot to climb the sloped terrain perfectly without vision, we need to adjust its orientation to make its body parallel to the walking surface. This is done by using the footstep location to estimate the slope of the ground. For each $i$-th leg, we can measure the foot position $\bm{p}_i = (p_{x,i}, p_{y,i}, p_{z,i})$ and build the vector of feet x-position ($\bm{p}_x$), y-position ($\bm{p}_y$), and z-position ($\bm{p}_z$). Thus, we can model the walking surface as a plane:
\begin{align}
z(x,y) = a_0 + a_1 x + a_2 y
\end{align}
and the coefficients ($a_0$, $a_1$, and $a_2$) will be obtained through the solution of the least square problem using $\bm{p}_x$, $\bm{p}_x$, and $\bm{p}_x$ data (see \cite{Bledt2018} for more details).

Note that the desired roll and pitch angles for the robot will be modified on the slope according to the following:
\begin{align}
\text{roll} = \arctan (a_2) , \quad \text{pitch} = \arctan(a_1).
\end{align}
As a result, the reference model's desired pitch and roll angles must be adjusted to the non-zero values determined as described above. It's important to note that the reference model utilizes the actual foot position of the robot, so there is no need to make any changes to the reference model's footstep planning when the robot is attempting to climb a slope.
\section{Conclusion} \label{sec: conclusion}
In conclusion, a novel control system has been presented that incorporates adaptive control into force control for legged robots walking under significant uncertainties. We have demonstrated the effectiveness of our proposed approach using numerical and experimental validations. The experiments show the success of the implementation of the proposed adaptive force control on quadruped robots, allowing them to walk and run while carrying an unknown heavy load on their trunk. The results are remarkable, with the robot being able to carry a load of up to 5 kg (50\% of its weight) while still keeping the tracking error within a small range and maintaining stability even in all directions. The experiment demonstrates that the proposed adaptive force control system cannot only adapt to model uncertainty but also leverage the benefits of force control in navigating rough terrains and soft terrain. On the other hand, the baseline non-adaptive controller fails to track the desired trajectory and causes the robot to collapse under uncertainty.

In the future, our goal is to broaden this methodology for auto-tuning MPC parameters, employing adaptive controllers specifically tailored for legged robot locomotion across various scenarios.

\section*{Acknowledgments}
This work is partially supported by the National Science Foundation Grant IIS-2133091 and the USC startup fund. The opinions expressed are those of the authors and do not necessarily reflect the opinions of the sponsors.
{\appendix

\subsection{Linear Quadratic Lyapunov Theory}\label{subsec:LQL}
According to Lyapunov theory \cite{Ames2014}, the PD control described in \eqref{eq: PDcontrol} will asymptotically stabilize the system if
\begin{equation}
\bm{A}_m = \begin{bmatrix}\bm{0}_6 & \bm{1}_6 \\-\bm{K}_P & -\bm{K}_D\end{bmatrix} \in \mathbb{R}^{12 \times 12}
\label{Amatrix}
\end{equation}
is Hurwitz.	This means that by choosing a control Lyapunov function candidate as follows:
\begin{equation}
\label{VUnscaledCoords}
V(\bm{e}) = \bm{e}^{T} \bm{P} \bm{e},
\end{equation}
where $\bm{P}\in \mathbb{R}^{12 \times 12}$ is the solution of the Lyapunov equation 
\begin{equation}\label{eq:Lyp}
{\bm{A}_m}^T \bm{P} + \bm{P} \bm{A}_m = -\bm{Q}_L, 
\end{equation}
and $\bm{Q}_L\in \mathbb{R}^{12 \times 12}$ is any symmetric positive-definite matrix. We then have:
\begin{align}\label{eq:LQL property}
\dot{V}(\bm{e},\bm{u}) + \lambda V(\bm{e}) = &~
\bm{e}^T ({\bm{D}_l}^T \bm{P} + \bm{P} {\bm{D}_l}) \bm{e} \nonumber\\ &+ \lambda V(\bm{e}) +2 \bm{e}^T \bm{P} {\bm{B}} \bm{u} ~\leq 0,
\end{align}
where,
\begin{align} 
\lambda = \frac{\lambda_{min}(\bm{Q}_L)}{\lambda_{max}(\bm{P})} > 0.
\end{align}
As a result, the state variable $\bm{e}$ and the control input $\bm{u}$ always remain bounded:
\begin{align}
\label{bounded eta}
\|\bm{e} \| \leq \delta_{\eta}, \quad \|\bm{u} \| \leq \delta_{u}.
\end{align}

However, the control signal $\bm{u}^*$ \eqref{eq: u_star} we construct by solving QP problem \eqref{eq: BalanceControlQP}, is not always the same as $\bm{u}$. Based on the friction constraints present in equation \eqref{eq: BalanceControlQP}, the value of $\bm{F}^*$ is always bounded. Besides, according to the definition of $\bm{A}$, $\bm{M}$, and $\bm{G}$, these matrices also have bounded values. Thus, it implies that:
\begin{align}
\label{bounded_mu}
\|\bm{u}^* \| \leq \delta_{{u^*}}.
\end{align}
Therefore, the vector of difference between $\bm{u}$ and $\bm{u}^*$ can be defined as:
\begin{align}\label{eq:mu_star}
\bm{\Delta} = \bm{u}^* - \bm{u}
\end{align}
which is also bounded according to \eqref{bounded_mu} and \eqref{bounded eta}:
\begin{align}\label{bounded_delta}
\|\bm{\Delta} \| \leq \delta_{\Delta}.
\end{align}
By substituting $\bm{u}^*$ in \eqref{eq:LQL property}, we have:
\begin{equation}\label{eq:LQL_mu_star}
\dot{V}(\bm{e},\bm{u}^*) + \lambda V(\bm{e}) \leq 2 \bm{e}^T \bm{P} {\bm{B}}\bm{\Delta} \leq \epsilon_{V},
\end{equation}
where
\begin{equation}\label{eq:epsilon_v}
\epsilon_{V} = 2 \|\bm{P}\| \delta_{\eta} \delta_{\Delta}.
\end{equation}

\subsection{Stability Analysis}
\textit{Theorem}: Consider the system dynamics with uncertainty described by \eqref{eq: EtaClosedLoopUncertainty}, and a reference model described by \eqref{eq: ref_model}. Assume the use of an $L_1$ adaptive controller with the optimal closed-loop control signal given by \eqref{eq: u_star}, the adaptive control signal given by \eqref{eq: controller mu2}, and the adaptation laws given by \eqref{eq: adap_law}. Then, under the aforementioned $L_1$ adaptive controller, the tracking error between the real model and reference model denoted as $\tilde{\bm{e}}$, as well as the errors between the real and estimated uncertainty, denoted as $\tilde{\bm{\alpha}}$ and $\tilde{\bm{\beta}}$, respectively, are bounded.

\textit{Proof}: Let us consider the following control Lyapunov candidate function:
\begin{align}
\label{CLF tilde}
\tilde{V}=\tilde{\bm{e}}^{T}\bm{P}\tilde{\bm{e}}+\tilde{\bm{\alpha}}^{T}\bm{\Gamma}^{-1}\tilde{\bm{\alpha}}+\tilde{\bm{\beta}}^{T}\bm{\Gamma}^{-1}\tilde{\bm{\beta}}.
\end{align}
Therefore, its time derivative will be
\begin{align}
\label{dotV}
\dot{\tilde{V}}=\dot{\tilde{\bm{e}}}^{T}\bm{P}\tilde{\bm{e}}+\tilde{\bm{e}}^{T}\bm{P}\dot{\tilde{\bm{e}}} +
\dot{\tilde{\bm{\alpha}}}^{T}\bm{\Gamma}^{-1}\tilde{\bm{\alpha}}+\tilde{\bm{\alpha}}^{T}\bm{\Gamma}^{-1}\dot{\tilde{\bm{\alpha}}}
\nonumber \\
+
\dot{\tilde{\bm{\beta}}}^{T}\bm{\Gamma}^{-1}\tilde{\bm{\beta}}+\tilde{\bm{\beta}}^{T}\bm{\Gamma}^{-1}\dot{\tilde{\bm{\beta}}}, 
\end{align}
in which we have
\begin{align}
\label{CLF eta parts}
\dot{\tilde{\bm{e}}}^{T}\bm{P}\tilde{\bm{e}}+\tilde{\bm{e}}^{T}\bm{P}\dot{\tilde{\bm{e}}} 
&=
({\bm{D}_l}\tilde{\bm{e}}+{\bm{B}}{\tilde{\bm{F}}})^{T}\bm{P}\tilde{\bm{e}}
+
\tilde{\bm{e}}^{T}\bm{P}({\bm{D}_l}\tilde{\bm{e}}+{\bm{B}}{\tilde{\bm{F}}}) \nonumber \\
&~~+
\tilde{\bm{\alpha}}^{T}{\bm{B}}^{T}||\bm{e}||\bm{P}\tilde{\bm{e}}+\tilde{\bm{e}}^{T}\bm{P}{\bm{B}}\tilde{\bm{\alpha}}||\bm{e}|| \nonumber \\
&~~
+\tilde{\bm{\beta}}^{T}{\bm{B}}^{T}\bm{P}\tilde{\bm{e}}+\tilde{\bm{e}}^{T}\bm{P}{\bm{B}}\tilde{\bm{\beta}}. 
\end{align}
Because $\tilde{\bm{e}}=\hat{\bm{e}}-\bm{e}$ satisfies the condition imposed by \eqref{eq:LQL_mu_star}, it implies that:
\begin{align}
\label{RES eta tilde}
&({\bm{D}_l}\tilde{\bm{e}}+{\bm{B}}\tilde{\bm{F}})^{T}\bm{P}\tilde{\bm{e}}
~+ \nonumber \\ &
\tilde{\bm{e}}^{T}\bm{P}({\bm{D}_l}\tilde{\bm{e}}+{\bm{B}}\tilde{\bm{F}})
\le 
-\lambda\tilde{\bm{e}}^{T}\bm{P}\tilde{\bm{e}} + \epsilon_{\tilde{V}},
\end{align}
where
\begin{equation}\label{eq:epsilon_v_tilde}
\epsilon_{\tilde{V}} = 2 \|\bm{P}\| \delta_{\tilde{e}} \delta_{\tilde{\Delta}}.
\end{equation}

Furthermore, with the property of the projection operator \cite{Lavretsky2011}, we have the following:
\begin{align}
\label{proj_property}
(\hat{\bm{\alpha}}-\bm{\alpha})^{T}(\text{Proj}(\hat{\bm{\alpha}},\bm{y}_{\alpha})-\bm{y}_{\alpha})\le 0, \nonumber \\
(\hat{\bm{\beta}}-\bm{\beta})^{T}(\text{Proj}(\hat{\bm{\beta}},\bm{y}_{\beta})-\bm{y}_{\beta})\le 0.
\end{align}
From \eqref{eq: adap_law} and \eqref{proj_property}, we can imply that
\begin{align}
\label{proj imply}
\tilde{\bm{\alpha}}^{T}\bm{\Gamma}^{-1}\dot{\tilde{\bm{\alpha}}} \le \tilde{\bm{\alpha}}^{T}\bm{y}_{\alpha}-\tilde{\bm{\alpha}}^{T}\bm{\Gamma}^{-1}\dot{\bm{\alpha}}, \nonumber \\ 
\tilde{\bm{\beta}}^{T}\bm{\Gamma}^{-1}\dot{\tilde{\bm{\beta}}} \le \tilde{\bm{\beta}}^{T}\bm{y}_{\beta}-\tilde{\bm{\beta}}^{T}\bm{\Gamma}^{-1}\dot{\bm{\beta}}.
\end{align}
We now replace \eqref{CLF eta parts}, \eqref{RES eta tilde} and \eqref{proj imply} to \eqref{dotV}, which results in
\begin{align}
\dot{\tilde{V}} & \le
-\lambda\tilde{\bm{e}}^{T}\bm{P}\tilde{\bm{e}} + \epsilon_{\tilde{V}} \nonumber \\
&+
\tilde{\bm{\alpha}}^{T}(\bm{y}_{\alpha}+{\bm{B}}^{T}\bm{P}\tilde{\bm{e}}||\bm{e}||)-\tilde{\bm{\alpha}}^{T}\bm{\Gamma}^{-1}\dot{\bm{\alpha}} \nonumber \\
&+
(\bm{y}_{\alpha}^{T}+\tilde{\bm{e}}^{T}\bm{P}{\bm{B}}||\bm{e}||)\tilde{\bm{\alpha}}-\dot{\bm{\alpha}}^{T}\bm{\Gamma}^{-1}\bm{\alpha}
\nonumber \\
&+
\tilde{\bm{\beta}}^{T}(\bm{y}_{\beta}+{\bm{B}}^{T}\bm{P}\tilde{\bm{e}})-\tilde{\bm{\beta}}^{T}\bm{\Gamma}^{-1}\dot{\bm{\beta}} \nonumber \\
&+
(\bm{y}_{\beta}^{T}+\tilde{\bm{e}}^{T}\bm{P}{\bm{B}})\tilde{\bm{\beta}}-\dot{\bm{\beta}}^{T}\bm{\Gamma}^{-1}\tilde{\bm{\beta}}
\end{align}
So, by using the chosen projection functions \eqref{eq:proj_fun}, then we conclude that:
\begin{align}
\label{RES_CLF bound}
\dot{\tilde{V}}+\lambda\tilde{V} \le  \epsilon_{\tilde{V}} +
\lambda\tilde{\bm{\alpha}}^{T}\bm{\Gamma}^{-1}\tilde{\bm{\alpha}}+
\lambda\tilde{\bm{\beta}}^{T}\bm{\Gamma}^{-1}\tilde{\bm{\beta}} \nonumber \\
-\tilde{\bm{\alpha}}^{T}\bm{\Gamma}^{-1}\dot{\bm{\alpha}} -\dot{\bm{\alpha}}^{T}\bm{\Gamma}^{-1}\tilde{\bm{\alpha}}
\nonumber \nonumber \\
-\tilde{\bm{\beta}}^{T}\bm{\Gamma}^{-1}\dot{\bm{\beta}} -\dot{\bm{\beta}}^{T}\bm{\Gamma}^{-1}\tilde{\bm{\beta}}.
\end{align}
We assume that the uncertainties $\bm{\alpha}$, $\bm{\beta}$, and their time derivatives are bounded. Furthermore, the projection operators \eqref{eq: adap_law} will also keep $\tilde{\bm{\alpha}}$ and $\tilde{\bm{\beta}}$ bounded (see \cite{L1_adaptive} for a detailed proof about these properties.) We define these bounds as follows:
\begin{align}
\label{bounded conditions for CLF proof}
||\tilde{\bm{\alpha}}|| \le& \tilde{\bm{\alpha}}_b ,~~
||\tilde{\bm{\beta}}|| \le \tilde{\bm{\beta}}_b , \nonumber \\
||\dot{\bm{\alpha}}|| \le& \dot{\bm{\alpha}}_b ,~~
||\dot{\bm{\beta}}|| \le \dot{\bm{\beta}}_b. 
\end{align}
Combining this with \eqref{RES_CLF bound}, we have,
\begin{align}
\label{relaxed CLF condition deltaV}
&\dot{\tilde{V}}+\lambda\tilde{V} \le \lambda\delta_{\tilde{V}}, 
\end{align}
where 
\begin{align}
\delta_{\tilde{V}}=2||\bm{\Gamma}||^{-1}(\tilde{\bm{\alpha}}_{b}^2+\tilde{\bm{\beta}}_{b}^{2}+\frac{1}{\lambda}\tilde{\bm{\alpha}}_{b}\dot{\bm{\alpha}}_{b}+\frac{1}{\lambda}\tilde{\bm{\beta}}_{b}\dot{\bm{\beta}}_{b}) + \frac{1}{\lambda}\epsilon_{\tilde{V}}.
\end{align}
Thus, if $\tilde{V} \ge \delta_{\tilde{V}}$ then $\dot{\tilde{V}}\le 0$. As a result, we always have $\tilde{V} \le \delta_{\tilde{V}}$. 
In other words, by choosing the adaptation gain $\bm{\Gamma}$ sufficiently large and $\bm{P}$ relatively small, we can limit the Control Lyapunov Function \eqref{CLF tilde} in an arbitrarily small neighborhood $\delta_{\tilde{V}}$ of the origin. 
According to \eqref{Amatrix} and \eqref{eq:Lyp}, achieving a small value for $\bm{P}$ depends on choosing a proper value for $\bm{K}_P$, $\bm{K}_D$, and $\bm{Q}_L$. Therefore, the value of PD gains affects the stability of the whole system.
Finally, the tracking errors between the dynamics model \eqref{eq: EtaClosedLoopUncertainty} and the reference model \eqref{eq: ref_model}, $\tilde{\bm{e}}$, and the error between the real and estimated uncertainty, $\tilde{\bm{\alpha}}$, $\tilde{\bm{\beta}}$ are bounded as follows:
\begin{align}
||\tilde{\bm{e}}|| \le \sqrt{\frac{\delta_{\tilde{V}}}{||\bm{P}||}} ,
||\tilde{\bm{\alpha}}|| \le \sqrt{||\bm{\Gamma}||\delta_{\tilde{V}}} ,||\tilde{\bm{\beta}}|| \le \sqrt{||\bm{\Gamma}||\delta_{\tilde{V}}}.
\end{align}}
\bibliography{references}

\begin{IEEEbiography}[{\includegraphics[width=1in,height=1.25in,clip,keepaspectratio]{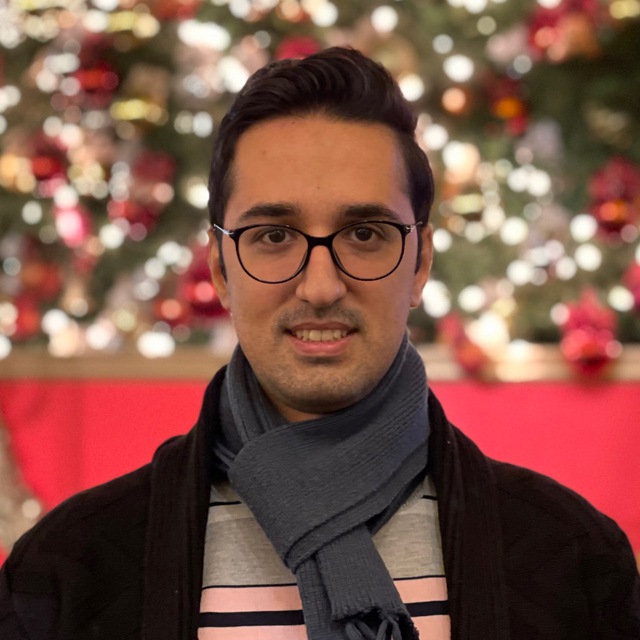}}]{Mohsen Sombolestan}
received his B.Sc. degree in mechanical engineering in 2017 from Sharif University of Technology, Tehran, Iran, and his M.Sc. degree in mechanical engineering in 2020 from Isfahan University of Technology, Isfahan, Iran. He is working toward a Ph.D. in mechanical engineering from University of Southern California, Los Angeles, CA, USA. 

His research interests include control system design in robotic applications, especially legged robots, focusing on adaptive control, model predictive control, and reinforcement learning.
\end{IEEEbiography}

\begin{IEEEbiography}[{\includegraphics[width=1in,height=1.25in,clip,keepaspectratio]{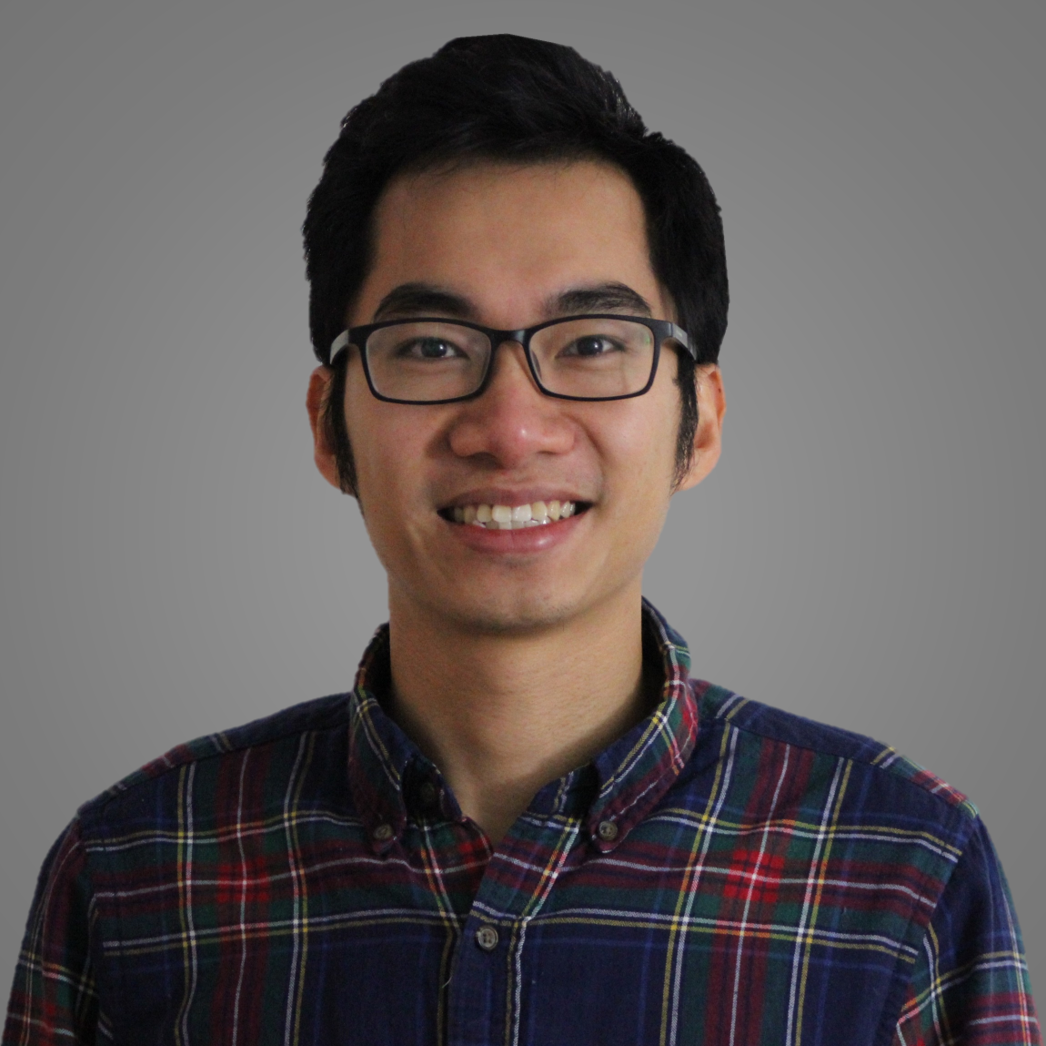}}]{Quan Nguyen}
is an assistant professor of Aerospace and Mechanical Engineering at the University of Southern California (USC). Before joining USC, he was a Postdoctoral Associate in the Biomimetic Robotics Lab at the Massachusetts Institute of Technology (MIT). He received his Ph.D. from Carnegie Mellon University (CMU) in 2017 with the Best Dissertation Award.
 
His research interests span different control and optimization approaches for highly dynamic robotics, including nonlinear control, trajectory optimization, real-time optimization-based control, and robust and adaptive control. His work on the MIT Cheetah 3 robot leaping on a desk was featured widely in many major media channels, including CNN, BBC, NBC, ABC, etc. Nguyen won the Best Presentation of the Session at the 2016 American Control Conference (ACC) and the Best System Paper Finalist at the 2017 Robotics: Science \& Systems Conference (RSS).
\end{IEEEbiography}

\end{document}